\documentclass[11pt]{article}

\PassOptionsToPackage{hyperfootnotes=false}{hyperref}

\usepackage[final]{acl}

\usepackage{times}
\usepackage{latexsym}
\usepackage[T1]{fontenc}
\usepackage[utf8]{inputenc}
\usepackage{microtype}
\usepackage{inconsolata}

\usepackage{booktabs}
\usepackage{amsfonts}
\usepackage{nicefrac}
\usepackage{amsmath}
\usepackage{amssymb}
\usepackage{algorithm}
\usepackage{algorithmic}
\usepackage{graphicx}
\usepackage{multirow}
\usepackage[table]{xcolor}
\usepackage{array}
\usepackage{hyperref}

\newcommand{\up}[1]{\textcolor[HTML]{008000}{\scriptsize (+#1)}}
\newcommand{\dn}[1]{\textcolor[HTML]{D32F2F}{\scriptsize (-#1)}}
\newcommand{\eq}{\textcolor[HTML]{000000}{\scriptsize (0.00)}}

\title{PACE: A Unified Condense-and-Extract Paradigm for Fast VLM Inference}

\author{
 Junjie Liu\textsuperscript{1} \quad
 Shengyuan Ye\textsuperscript{2} \quad
 Xu Chen\textsuperscript{1,3}\thanks{\ Corresponding author: \href{mailto:chenxu35@mail.sysu.edu.cn}{\texttt{chenxu35@mail.sysu.edu.cn}}} \\
 \textsuperscript{1}Sun Yat-sen University \\
 \textsuperscript{2}Power Dispatch Control Center, Guangdong Power Grid Co., Ltd. \\
 Guangzhou, China \\
 \textsuperscript{3}Shenzhen Loop Area Institute, Shenzhen, China
}

\begin{document}

\maketitle

\begin{abstract}
Vision-Language Models (VLMs) demonstrate exceptional visual reasoning capabilities, yet their inference costs escalate rapidly with the proliferation of visual tokens. Existing visual token pruning methods exhibit two fundamental limitations. First, most approaches operate exclusively post-vision encoder, leaving the substantial latency of the visual encoding phase unoptimized. Second, under strict token budgets, these methods often fail to jointly preserve holistic visual contexts and fine-grained details, leading to performance degradation. To address these bottlenecks, we propose \textbf{PACE} (Pixel-Adaptive Condense and Extract), a training-free inference framework that accelerates both the vision encoder and the Large Language Model (LLM) via a unified \textit{Condense-and-Extract} paradigm. During the \textit{Condense} stage, an Adaptive Pixel Compressor (APC) evaluates visual information density prior to encoding, adaptively downsampling redundant inputs, curtailing encoder computation while preserving global context and essential visual cues. In the \textit{Extract} stage, a Dynamic Dual-Attention Extractor (DDAE) selectively retains visual tokens via a fusion of internal visual signals from the encoder and semantic signals from the LLM, safeguarding task-critical details. By integrating PACE into Qwen2.5-VL-7B, the model retains 93.8\% of its original performance while utilizing only 10\% of the visual tokens, yielding a $3.1\times$ speedup in time to first token (TTFT). Our code is available at \url{https://github.com/jjL357/PACE}.
\end{abstract}

\begin{figure}[t]
 \centering
 \includegraphics[width=\columnwidth]{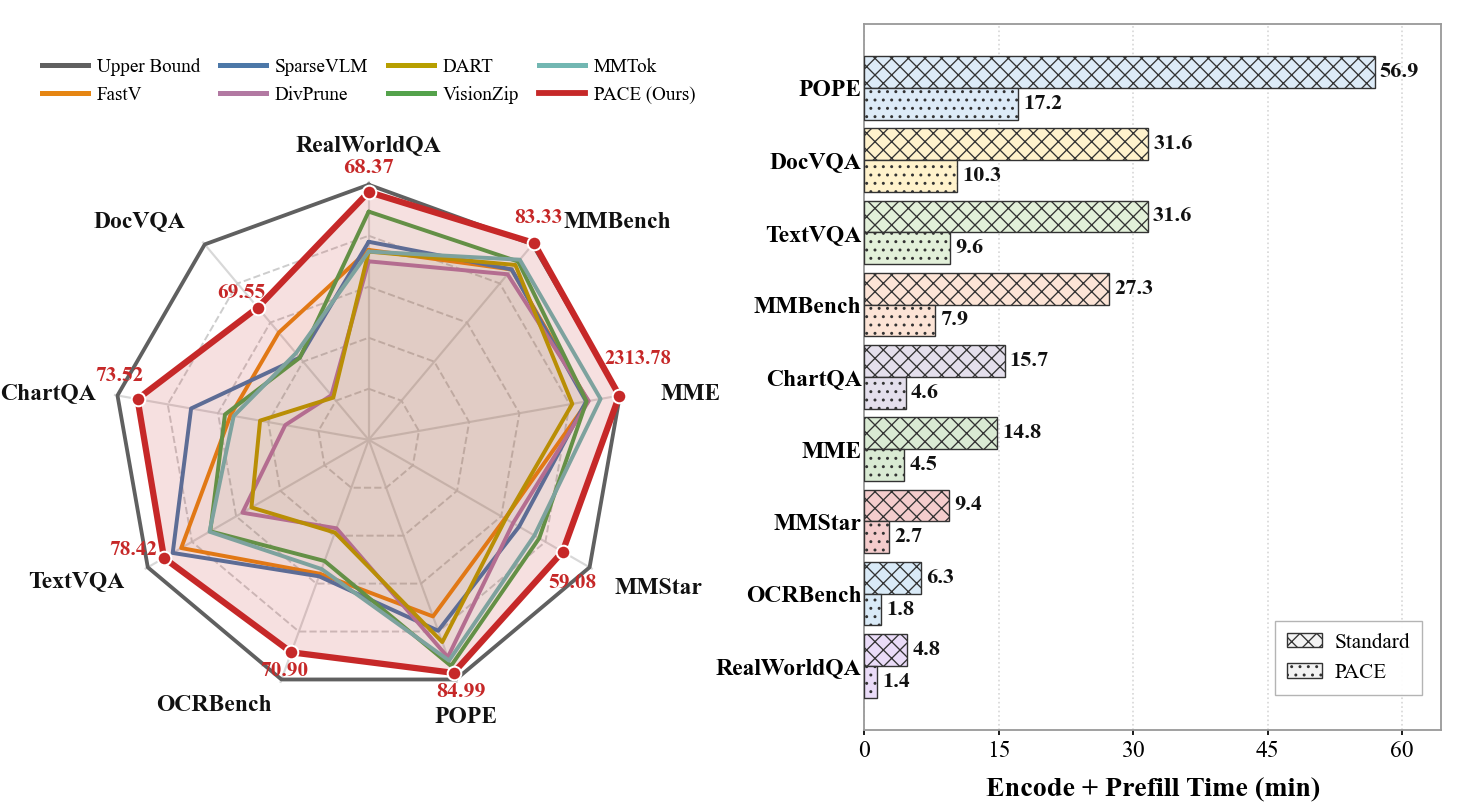}
 \caption{\textbf{Accuracy--TTFT trade-off at 10\% visual-token retention.}
 PACE preserves accuracy while reducing pre-generation latency on Qwen2.5-VL-7B; standard denotes Vanilla inference of the original model.}
 \label{fig:pace_performance}
\end{figure}

\section{Introduction}

Vision-Language Models (VLMs) have established a new standard for multimodal understanding by seamlessly integrating visual perception with linguistic reasoning \citep{alayrac2022flamingo, liu2023visual}. Recent architectures have extended these capabilities to highly complex reasoning tasks \citep{dai2023instructblip,team2023gemini, li2025mini}. However, this rapid advancement demands an ever-expanding visual token budget. Processing high-resolution images \citep{achiam2023gpt, liu2024llavanext} or extensive videos \citep{liu2025video, wang2025accelerating} generates massive visual token sequences. Furthermore, modern VLMs equipped with native dynamic resolution capabilities \citep{bai2023qwen, guo2025seed1, abouelenin2025phi} partition inputs into numerous patches, frequently producing substantial visual redundancy. For instance, encoding a 4K image in Qwen2.5-VL generates over 42,000 patches for the Vision Transformer (ViT) \citep{dosovitskiy2020image}; even after spatial pooling, over 10,500 visual tokens extend the LLM context. Given the quadratic computational complexity of self-attention \citep{dao2022flashattention}, this proliferation imposes a severe inference bottleneck, hindering real-time deployment.

Compressing the visual token sequence is therefore imperative for efficient inference. Visual token pruning naturally addresses this challenge by identifying and retaining critical visual tokens while discarding redundant ones prior to LLM decoding. Current pruning methods discard tokens based on attention distributions \citep{chen2024image,takezoe2026learnpruner}, token similarity \citep{wen2025stop,zou2026don}, diversity \citep{zhang2026beyond,fang2026prune}, or maximum coverage \citep{dong2025mmtok,deng2026scope}. While these strategies effectively alleviate LLM-side computational demands, they share two fundamental limitations.

\begin{figure}[t] 
 \centering
 \includegraphics[width=\columnwidth]{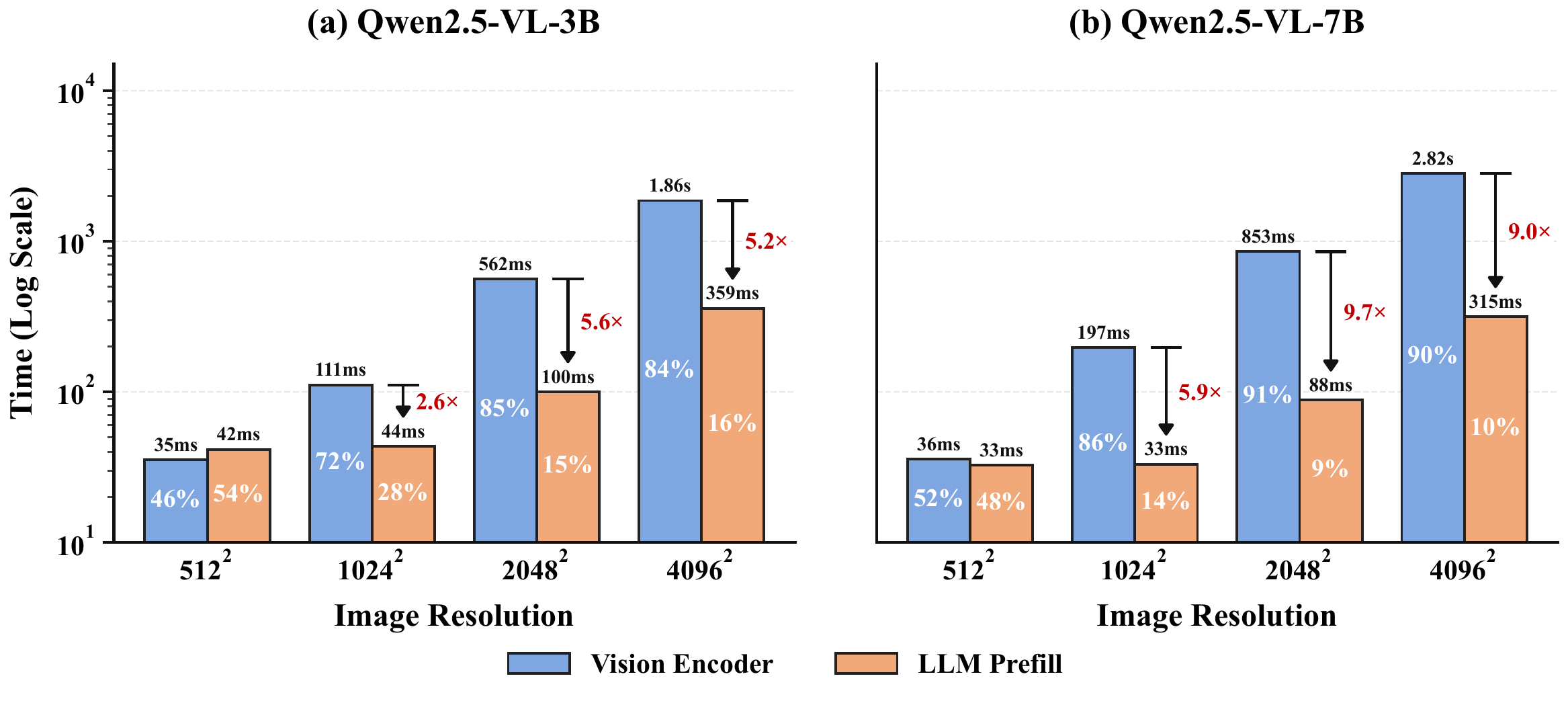} 
 \caption{\textbf{Latency decomposition for Qwen2.5-VL.}
 Vision encoding and LLM prefill jointly dominate latency as input resolution increases.}
 \label{fig:qwen_profiling_results}
\end{figure}

\paragraph{The Dual Bottleneck of High-Resolution Inference.}
Prevailing visual token pruning methods operate exclusively \textit{after} the vision encoder. They truncate the LLM context sequence but neglect the substantial computational overhead of the vision encoder itself. Empirical profiling reveals that at high resolutions, both the ViT encoding and LLM prefill stages impose severe latency bottlenecks, as illustrated in Figure~\ref{fig:qwen_profiling_results}. Because most visual token pruning methods intervene strictly during modality alignment or within the LLM layers, the immense computational cost of encoding high-resolution pixels remains unaddressed. Consequently, truncating only the LLM-side sequence resolves merely half of the inference bottleneck.

\begin{figure}[t]
 \centering
 \includegraphics[width=\columnwidth]{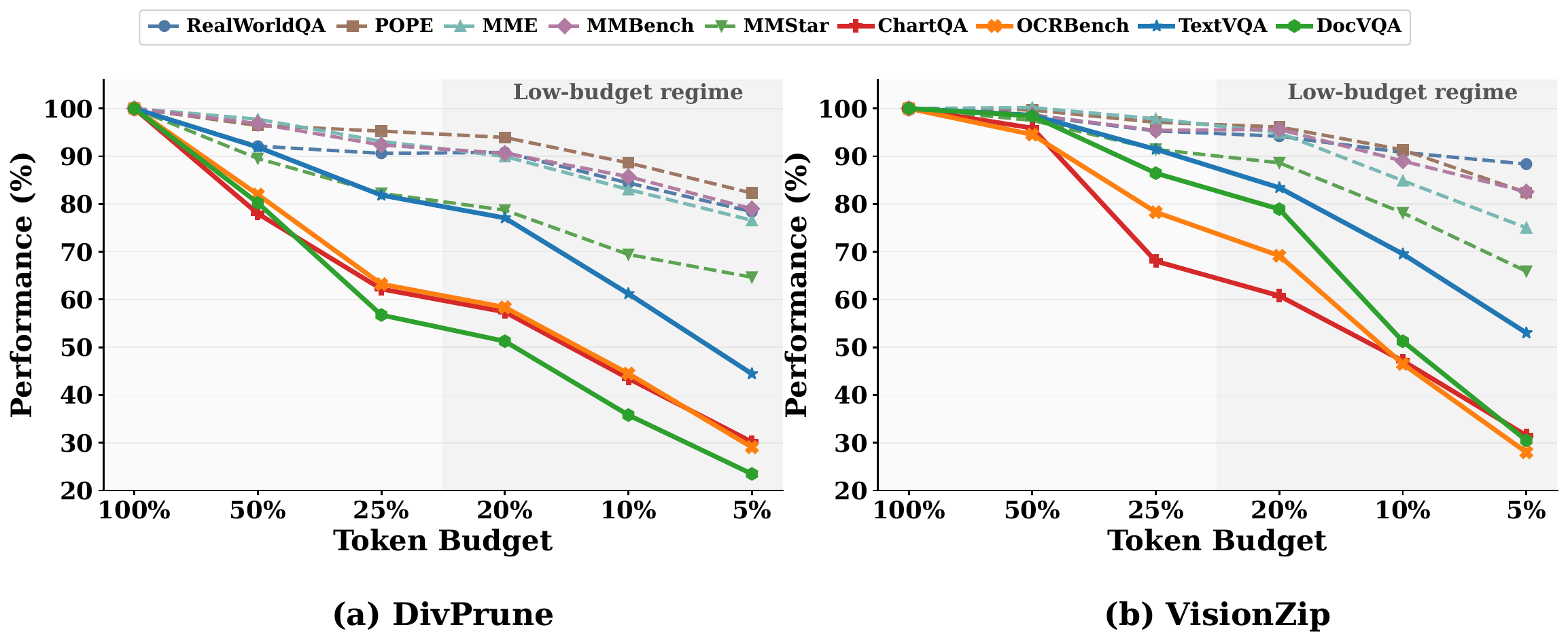}
 \caption{
 \textbf{Performance of existing visual token pruning methods under varying token budgets on Qwen2.5-VL-7B.}
 Performance on detail-sensitive benchmarks degrades sharply under aggressive token pruning.}
 \label{fig:dynamic_res_budget_collapse}
\end{figure}

\paragraph{Information and Detail Loss under Low-Budget Compression.}
Furthermore, under strict token budgets, existing methods struggle to retain holistic visual contexts and fine-grained details simultaneously. Aggressive pruning inherently fragments visual layouts and discards indispensable visual cues, such as text strokes and alignment anchors. This structural loss results in severe performance degradation, particularly on detail-sensitive tasks. While competitive methods, including DivPrune and VisionZip, maintain robust accuracy on general benchmarks at a 5\% token budget, they experience severe performance drops on detail-sensitive tasks such as ChartQA and DocVQA (Figure~\ref{fig:dynamic_res_budget_collapse}). This vulnerability indicates that simple post-encoder token pruning fails to preserve the critical cues essential for high-density visual reasoning.

These limitations underscore the necessity for a paradigm shift: visual token compression must transcend naive post-encoder sequence reduction by simultaneously preserving holistic layouts and salient visual details \textit{prior} to intensive computation. An optimal framework must fulfill two complementary functions. First, it should \textit{condense} visual information before the expensive encoding phase. Recent studies \citep{ye2025voco,cai2025matryoshka} suggest that condensed visual representations inherently carry concentrated semantic meaning. Pre-encoder condensation ensures the ViT operates on a compact pixel budget without sacrificing layout integrity. Second, it should \textit{extract} the most informative visual tokens post-encoding, guaranteeing that critical, task-relevant details are prioritized within the LLM context.

To this end, we propose \textbf{PACE}, a training-free, plug-and-play inference framework organized around a unified \textbf{Condense-and-Extract} paradigm. In the \textbf{Condense} stage, the Adaptive Pixel Compressor (APC) uses a lightweight feature preview to estimate information density. It dynamically condenses the input image prior to the vision encoder, globally preserving visual cues under strict pixel budgets. In the \textbf{Extract} stage, the Dynamic Dual-Attention Extractor (DDAE) fuses ViT self-attention with LLM cross-modal attention. This dynamic dual-attention mechanism enables PACE to extract fine-grained details reliably, markedly outperforming single-source attention pruning. Figure~\ref{fig:pace_performance} summarizes the resulting accuracy--TTFT trade-off.

Our contributions are summarized as follows:
\begin{itemize}
 \item \textbf{Identification of the Dual Bottleneck.} We demonstrate that conventional visual token pruning overlooks the substantial vision-encoder overhead and incurs severe degradation on detail-sensitive tasks due to its inability to jointly preserve holistic contexts and fine-grained details.
 \item \textbf{Unified Condense-and-Extract Framework.} We introduce PACE, seamlessly integrating pre-encoder adaptive pixel condensation (APC) with post-encoder dynamic dual-attention extraction (DDAE). APC constructs a compact visual representation, while DDAE safeguards salient visual tokens via confidence-weighted attention fusion.
 \item \textbf{Superior Performance--Efficiency Trade-off.} Extensive evaluations demonstrate that PACE consistently outperforms existing pruning baselines. When deployed on Qwen2.5-VL-7B, PACE preserves over 93\% of the original performance while discarding 90\% of the visual tokens, unlocking a $3.1\times$ TTFT acceleration.
\end{itemize}

\section{Related Work}
\label{sec:related_work}

\subsection{Vision-Language Models}
Early VLMs typically rely on fixed-resolution vision encoders, requiring images to be resized or padded before visual encoding \citep{liu2023visual}. This design simplifies model processing but may lose fine-grained details. To improve high-resolution perception, models such as InternVL \citep{chen2024internvl} adopt dynamic high-resolution tiling, where images are split into multiple local crops according to their aspect ratio and resolution. More recent models, including Qwen2.5-VL \citep{bai2025qwen25vltechnicalreport} and Qwen3-VL \citep{bai2025qwen3}, support native dynamic-resolution processing, preserving image aspect ratios and producing variable-length visual token sequences. However, as the visual token length still grows with the input pixel budget, high-resolution inference remains computationally expensive \citep{dao2024flashattention}.

\subsection{Visual Token Pruning}
Visual token pruning reduces inference cost by selecting or merging a subset of visual tokens before they enter the LLM. Attention-based methods, such as FastV \citep{chen2024image} and SparseVLM \citep{zhang2024sparsevlm}, estimate token importance from attention or vision--language relevance. Hybrid methods, including LLaVA-PruMerge \citep{shang2025llava} and VisionZip \citep{yang2025visionzip}, combine importance-based selection with similarity-based merging. Other methods exploit redundancy, diversity, or coverage, such as DART \citep{wen2025stop}, DivPrune \citep{alvar2025divprune}, and MMTok \citep{dong2025mmtok}. Although these approaches effectively reduce LLM-side prefill cost, they are mostly applied after visual encoding. Thus, the ViT encoding cost remains unchanged.

\subsection{Adaptive Resolution VLMs}
Adaptive-resolution VLMs adjust image scale, compression rate, or token allocation according to visual complexity. Existing methods often require additional modules, learned routing, or reinforcement-learning procedures. For example, ViCO \citep{cui2025vico} and HyperVL \citep{team2025hypervl} introduce learned compression or routing mechanisms, while AdaptVision \citep{lin2025adaptvision} and VisionThink \citep{yang2026visionthink} formulate adaptive resolution as coarse-to-fine visual acquisition. In contrast, PACE is training-free and requires no architectural modification. It reduces both pre-encoder and post-encoder costs by adapting input resolution before visual encoding and controlling the visual tokens passed to the LLM, while preserving global layout and fine-grained visual evidence.

\section{Method}
\label{sec:method}

\subsection{Overview of PACE}

\begin{figure*}[t]
 \centering
 \includegraphics[width=0.95\textwidth]{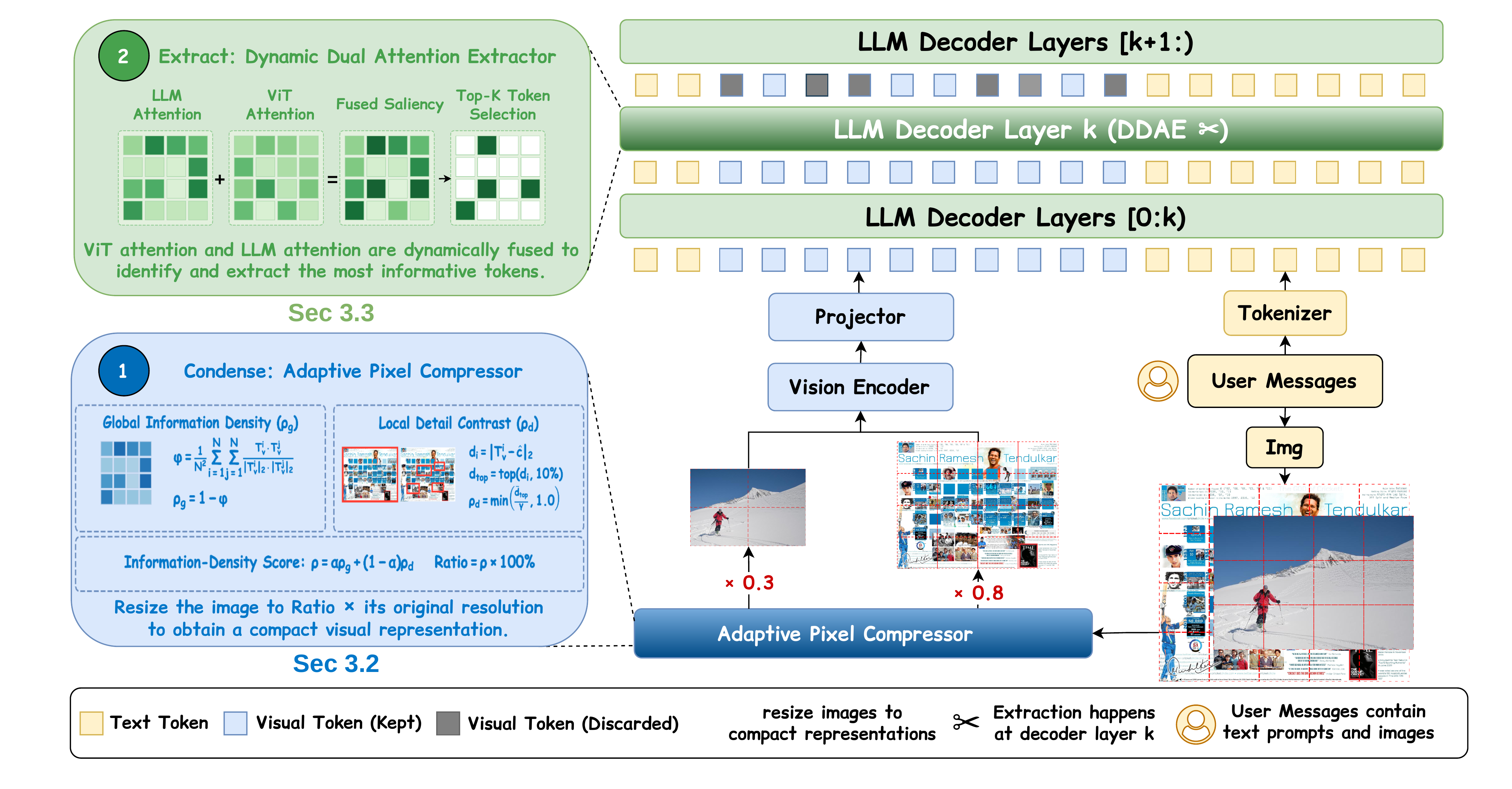}
 \caption{
 \textbf{Overview of PACE's unified Condense-and-Extract pipeline.}
 In the Condense stage, APC uses a shallow visual preview to estimate global redundancy and local detail, then adaptively resizes the input before the vision encoder. In the Extract stage, DDAE fuses semantic attention from the LLM with self-attention from the ViT and retains the top-$K$ visual tokens at decoder layer $k$. Together, APC lowers vision-encoder cost and DDAE shortens LLM prefill, enabling efficient low-budget inference while preserving holistic layouts and fine-grained task evidence.
 }
 \label{fig:overview}
\end{figure*}

As illustrated in Figure~\ref{fig:overview}, PACE introduces a unified \textit{Condense-and-Extract} pipeline designed to overcome the dual computational bottlenecks of the ViT and LLM, as well as the severe detail loss typically induced by low-budget token pruning. In the \textbf{Condense} stage, PACE evaluates visual information density prior to the ViT forward pass. Using a shallow feature preview to quantify detail complexity, PACE adaptively condenses the input image. This mechanism strictly curtails the pixel volume processed by the vision encoder while maintaining essential layout cues. In the \textbf{Extract} stage, PACE eliminates residual visual redundancy post-encoding. It dynamically integrates cross-attention from the LLM with self-attention from the ViT, ensuring that token selection prioritizes both semantic relevance and fine-grained detail integrity.

\subsection{Condense: Adaptive Pixel Compressor}
\label{sec:apc}

While heuristically dropping uninformative patches (such as uniform backgrounds or regions lacking salient information) \citep{wang2026pixelprune, choi2026docprune} intuitively reduces input length, it irrevocably disrupts the continuous 2D topological layout required by modern dynamic-resolution ViTs, risking the deletion of latent contextual anchors. Consequently, although heuristic dropping may suffice for narrow, domain-specific applications, it generalizes poorly to diverse, open-world visual scenarios. Conversely, static uniform downsampling blurs microscopic elements, such as text strokes, while continuing to allocate wasteful computation to homogeneous backgrounds.

To address this dilemma, we introduce the Adaptive Pixel Compressor (APC) in the \textbf{Condense} stage. Instead of pruning raw patches, APC dynamically modulates the global input resolution, allocating higher pixel budgets to information-dense inputs and lower budgets to redundant ones. This approach preserves continuous visual layouts while dynamically adapting to image complexity.

\paragraph{Shallow Feature Preview.} 
Instead of relying on basic pixel-level statistics, APC uses the initial ViT block ($K=1$) to compute a lightweight feature preview, similar to AdaPatch \citep{liuone}. This minimizes preprocessing overhead while securing essential semantic priors. The preview yields token embeddings, which are subsequently $\ell_2$-normalized and denoted as $\mathbf{T}_v = \{\mathbf{T}_v^i\}_{i=1}^N \in \mathbb{R}^{N \times D}$. A controlled ablation in Appendix~\ref{sec:preview_ablation} confirms that this semantic preview is more reliable than RGB, entropy, edge-density, and Laplacian statistics.

\paragraph{Global Information Density ($\rho_g$).} 
We quantify global redundancy by calculating the average pairwise cosine similarity $\varphi$ across all visual tokens:
\begin{equation}
 \label{eq:avg_sim}
 \varphi = \frac{1}{N^2} \sum_{i=1}^N \sum_{j=1}^N \frac{\mathbf{T}_v^i \cdot \mathbf{T}_v^j}{\|\mathbf{T}_v^i\|_2 \cdot \|\mathbf{T}_v^j\|_2}.
\end{equation}
A higher $\varphi$ indicates pronounced redundancy (e.g., large uniform areas) and lower global information density. The global information density score $\rho_g$ is defined as the non-redundant fraction:
\begin{equation}
 \label{eq:rho_g}
 \rho_g = 1.0 - \varphi
\end{equation}

\paragraph{Local Detail Contrast ($\rho_d$).} 
Images dominated by uniform backgrounds may yield high global redundancy metrics while harboring sparse yet indispensable details, such as microscopic text on a large white document. To prevent the erasure of such cues, APC computes a global background baseline $\mathbf{c}$ by averaging all normalized tokens:
\begin{equation}
 \mathbf{c} = \frac{1}{N} \sum_{i=1}^N \mathbf{T}_v^i.
\end{equation}
This baseline is normalized as $\hat{\mathbf{c}} = \mathbf{c} / \|\mathbf{c}\|_2$. We then compute the Euclidean distance $d_i$ between each token and the reference baseline:
\begin{equation}
 d_i = \left\| \mathbf{T}_v^i - \hat{\mathbf{c}} \right\|_2.
\end{equation}
Tokens diverging significantly from the baseline correlate strongly with sharp local details. APC isolates the top 10\% of tokens with the largest $d_i$ and averages their distances to compute $\bar{d}_{top}$. This tail mean avoids diluting sparse details through a full-image average while being less noise-sensitive than a single maximum; 5\%--20\% tail choices behave similarly (Appendix~\ref{sec:tail_percentile}). This metric is linearly scaled into a local retention score:
\begin{equation}
 \rho_d = \min\left(\frac{\bar{d}_{top}}{\gamma}, \, 1.0\right),
\end{equation}
where $\gamma$ acts as a regulating scaling factor. A high $\rho_d$ signals sharp local contrast, mandating a higher target resolution to safeguard fine-grained visual features.

\paragraph{Adaptive Condensation.}
Finally, APC integrates the global and local scores into a unified target retention ratio $\rho$:
\begin{equation}
 \rho = \alpha \rho_g + (1 - \alpha) \rho_d,
\end{equation}
where $\alpha$ is a weighting hyperparameter. The target retention ratio is defined as $r = \rho$, and the raw input image is resized accordingly. Importantly, to ensure strict adherence to system memory constraints, if the allocated token budget specifies a maximum scaling ratio lower than $\rho$, the input is directly resized to satisfy the hard budget. Operating as an independent pre-encoder module, APC seamlessly integrates into other visual token pruning frameworks, as detailed in Section~\ref{sec:apc_compatibility}.

\subsection{Extract: Dynamic Dual-Attention Extractor}
\label{sec:ddae}

Although APC reduces the pre-encoder sequence length, encoded features may still harbor latent visual redundancy. The \textbf{Extract} stage removes this residual redundancy prior to the LLM prefill phase, isolating tokens that are both semantically relevant and informatively rich.

Conventional post-encoder extraction mechanisms depend heavily on LLM cross-attention. While this mapping accurately isolates prompt-related semantics, it can focus too narrowly on explicitly referenced regions: explicit textual references receive high attention, whereas critical visual anchors, such as chart grid lines, receive negligible weights and are often erroneously pruned. Conversely, ViT self-attention reliably demarcates visual boundaries but operates unconditioned on the query, frequently retaining task-irrelevant background clutter. Relying on either signal in isolation fails to capture both salient targets and fine-grained supporting details.

To resolve this single-modality bias, we introduce the \textbf{Dynamic Dual-Attention Extractor (DDAE)}, which adaptively fuses linguistic semantic signals and visual signals via confidence-weighted attention integration.

Initially, DDAE extracts semantic attention scores from a designated LLM layer (denoted as extraction depth $L_{ext}$) to formulate a semantic relevance map $S_{llm}$, min-max normalized to $[0,1]$. Concurrently, internal self-attention maps are fetched from the terminal layers of the vision encoder to construct a visual density map $S_{vis}$, similarly normalized to $[0,1]$.

DDAE employs the standard deviations, $\sigma_{llm}$ and $\sigma_{vis}$, of these normalized distributions as unsupervised confidence proxies. A higher standard deviation indicates a sharper distribution, signifying elevated confidence toward a concise set of salient regions. A softmax function dynamically assigns the fusion weights $\alpha_{weight}$ and $\beta_{weight}$:
\begin{equation}
 \alpha_{weight}, \beta_{weight} = \text{Softmax}\left(\frac{[\sigma_{llm}, \sigma_{vis}]}{\tau}\right)
\end{equation}
where $\tau$ serves as a tunable temperature parameter. The final token saliency score is aggregated as $S_{final} = \alpha_{weight} S_{llm} + \beta_{weight} S_{vis}$. DDAE ranks the visual tokens based on $S_{final}$ and preserves the top-$K$ tokens to formulate the final LLM context sequence. Detailed theoretical analysis regarding computational complexity reduction is provided in Appendix~\ref{sec:detailed_complexity}.

\begin{table*}[t]
\centering
\small
\renewcommand{\arraystretch}{1.1}
\resizebox{\textwidth}{!}{%
\begin{tabular}{l|ccccccccc|c}
\toprule
\textbf{Method} & \textbf{RealWorldQA} & \textbf{POPE} & \textbf{MME} & \textbf{MMBench} & \textbf{MMStar} & \textbf{ChartQA}& \textbf{OCRBench}& \textbf{TextVQA} & \textbf{DocVQA} & \textbf{Avg.} \\
& Acc. $\uparrow$ & F1 $\uparrow$ & P+C $\uparrow$ & Acc. $\uparrow$ & Acc. $\uparrow$ & Acc. $\uparrow$ & Acc. $\uparrow$ & Acc. $\uparrow$ & ANLS $\uparrow$ & $\uparrow$ \\
\midrule
\rowcolor[HTML]{EFEFEF} \multicolumn{11}{c}{\textit{Fixed-resolution setting (MinPix = $2048 \times 28 \times 28$, MaxPix = $2048 \times 28 \times 28$)}} \\
Vanilla (100\%) & 69.54 & 86.36 & 2317 & 82.99 & 63.91 & 78.20 & 77.30 & 82.37 & 94.74 & 100.0\% \\
\midrule
\rowcolor[HTML]{FAFAFA} \multicolumn{11}{c}{\textit{Retain 20\% $\bar{T}$} ($\downarrow$ \textbf{80\% Tokens})} \\
FastV (ECCV'24) & 64.58 & 80.99 & 2256 & 80.76 & 54.79 & 64.12 & 66.60 & 79.16 & 76.58 & 90.2\% \\
SparseVLM (ICML'25) & 66.41 & 83.23 & 2258 & 81.36 & 55.65 & 70.00 & 56.54 & 80.33 & 73.56 & 90.3\% \\
DivPrune (CVPR'25) & 61.83 & 84.07 & 2248 & 80.07 & 54.66 & 51.56 & 51.50 & 69.99 & 49.40 & 81.7\% \\
DART (EMNLP'25) & 63.53 & 82.81 & 2278 & 79.64 & 55.83 & 58.72 & 54.00 & 69.48 & 50.58 & 83.5\% \\
 VisionZip (CVPR'25) & \textbf{67.06} & 85.50 & 2317 & 81.01 & 58.93 & 69.68 & 64.60 & 77.52 & 75.77 & 92.5\% \\
MMTok (ICLR'26) & 63.79 & 84.81 & 2278 & 82.22 & 58.17 & 68.40 & 65.80 & 76.78 & 74.25 & 91.4\% \\
PACE (w/o APC) & 66.93 & 85.29 & 2310 & 81.36 & 59.29 & 72.84 & 69.10 & 80.72 & 83.00 & 94.9\% \\
 \rowcolor[HTML]{E8F5E9} \textbf{PACE (Ours)} & 66.93 & \textbf{86.23} & \textbf{2322} & \textbf{83.08} & \textbf{62.47} & \textbf{78.56} & \textbf{79.00} & \textbf{81.66} & \textbf{86.84} & \textbf{98.6\%} \\
\midrule
\rowcolor[HTML]{FAFAFA} \multicolumn{11}{c}{\textit{Retain 10\% $\bar{T}$} ($\downarrow$ \textbf{90\% Tokens})} \\
FastV (ECCV'24) & 59.08 & 73.28 & 2154 & 77.23 & 48.78 & 51.80 & 52.40 & 74.23 & 59.56 & 79.9\% \\
SparseVLM (ICML'25) & 60.39 & 76.25 & 2145 & 77.23 & 51.30 & 61.00 & 53.00 & 76.30 & 49.61 & 81.4\% \\
DivPrune (CVPR'25) & 57.25 & 81.73 & 2158 & 76.12 & 50.22 & 39.00 & 41.70 & 59.32 & 34.58 & 72.5\% \\
DART (EMNLP'25) & 58.82 & 78.60 & 2074 & 78.26 & 48.89 & 44.88 & 42.70 & 57.12 & 33.65 & 72.6\% \\
VisionZip (CVPR'25) & 65.23 & 83.55 & 2147 & 79.04 & 54.90 & 53.08 & 49.40 & 67.16 & 49.55 & 81.1\% \\
MMTok (ICLR'26) & 58.82 & 82.44 & 2218 & 79.47 & 54.04 & 51.04 & 51.30 & 67.33 & 51.37 & 80.4\% \\
PACE (w/o APC) & 65.36 & 82.27 & 2228 & 80.07 & 55.25 & 65.76 & 58.00 & 76.18 & 65.54 & 87.7\% \\
\rowcolor[HTML]{E8F5E9} \textbf{PACE (Ours)} & \textbf{68.37} & \textbf{84.99} & \textbf{2314} & \textbf{83.33} & \textbf{59.08} & \textbf{73.52} & \textbf{70.90} & \textbf{78.42} & \textbf{69.55} & \textbf{93.8\%} \\
\midrule
\rowcolor[HTML]{FAFAFA} \multicolumn{11}{c}{\textit{Retain 5\% $\bar{T}$} ($\downarrow$ \textbf{95\% Tokens})} \\
FastV (ECCV'24) & 56.08 & 62.45 & 1995 & 73.54 & 45.88 & 36.52 & 41.40 & 66.21 & 44.17 & 69.6\% \\
SparseVLM (ICML'25) & 56.60 & 65.25 & 2011 & 72.08 & 45.01 & 46.24 & 43.10 & 69.53 & 30.99 & 70.3\% \\
DivPrune (CVPR'25) & 53.46 & 77.66 & 1926 & 72.77 & 45.20 & 28.96 & 29.80 & 41.28 & 22.64 & 62.0\% \\
DART (EMNLP'25) & 52.68 & 71.33 & 1946 & 73.97 & 42.80 & 30.76 & 32.90 & 45.01 & 23.37 & 62.2\% \\
VisionZip (CVPR'25) & 61.57 & 78.65 & 2029 & 75.17 & 48.63 & 40.96 & 38.30 & 55.47 & 29.40 & 70.5\% \\
MMTok (ICLR'26) & 54.12 & 78.70 & 2133 & 75.17 & 47.00 & 32.76 & 36.00 & 50.95 & 28.85 & 67.3\% \\
PACE (w/o APC) & 60.39 & 76.59 & 2117 & 75.95 & 49.26 & 53.00 & 46.10 & 67.32 & 45.38 & 76.9\% \\
\rowcolor[HTML]{E8F5E9} \textbf{PACE (Ours)} & \textbf{63.79} & \textbf{80.70} & \textbf{2245} & \textbf{80.07} & \textbf{56.02} & \textbf{61.84} & \textbf{57.80} & \textbf{71.56} & \textbf{48.94} & \textbf{84.3\%} \\
\bottomrule
\end{tabular}%
}
\vspace{2mm}
\caption{\textbf{Performance comparison on Qwen2.5-VL-7B under the fixed-resolution setting.}
 Benchmark scores cover nine tasks at 20\%, 10\%, and 5\% visual-token retention. PACE delivers the strongest overall result at every evaluated budget, with particularly robust performance on detail-sensitive benchmarks such as OCRBench and DocVQA.}
\label{tab:main_results}
\end{table*}

\section{Experiments}
\label{sec:experiments}

\subsection{Evaluation Setup}

\paragraph{Models and Framework.}
We evaluate PACE on the Qwen2.5-VL architectures (3B and 7B variants) \citep{bai2025qwen25vltechnicalreport}. To ensure standardized and reproducible evaluations, we conduct all benchmark experiments using the open-source \textbf{\texttt{lmms-eval}} framework \citep{zhang2025lmms}. To assess cross-model generalization, we further evaluate PACE on InternVL3.5-4B \citep{wang2025internvl3_5}, whose multi-tile pipeline differs from Qwen's native dynamic-resolution design; the results are reported in Appendix~\ref{sec:internvl_generalization}.

\paragraph{Baselines.}
We benchmark PACE against prominent visual token pruning methods: \textbf{FastV} \citep{chen2024image}, \textbf{SparseVLM} \citep{zhang2024sparsevlm}, \textbf{DivPrune} \citep{alvar2025divprune}, \textbf{DART} \citep{wen2025stop}, \textbf{VisionZip} \citep{yang2025visionzip}, and \textbf{MMTok} \citep{dong2025mmtok}. 

\paragraph{Implementation Details.}
For the APC module, we set the preview depth $K=1$, fusion weight $\alpha = 0.6$, and local contrast scaling $\gamma = 1.5$. For DDAE, semantic maps are extracted from the second LLM layer ($L_{ext}=2$) with a fusion temperature $\tau=0.5$. Evaluations encompass two configurations: the \textbf{fixed-resolution setting} and the \textbf{dynamic-resolution setting}, with the former serving as the default unless otherwise specified. Detailed configurations are provided in Appendix~\ref{sec:appendix_setup}.

\paragraph{Benchmarks.}
Our evaluation encompasses nine distinct datasets: \textbf{MME} \citep{fu2026mme}, \textbf{POPE} \citep{li2023evaluating}, \textbf{MMBench} \citep{liu2024mmbench}, \textbf{MMStar} \citep{chen2024we}, \textbf{RealWorldQA} \citep{grok15}, \textbf{TextVQA} \citep{singh2019towards}, \textbf{DocVQA} \citep{mathew2021docvqa}, \textbf{ChartQA} \citep{masry2022chartqa}, and \textbf{OCRBench} \citep{liu2024ocrbench}.

\subsection{Main Results}

\begin{figure*}[t]
 \centering
 \includegraphics[width=\textwidth]{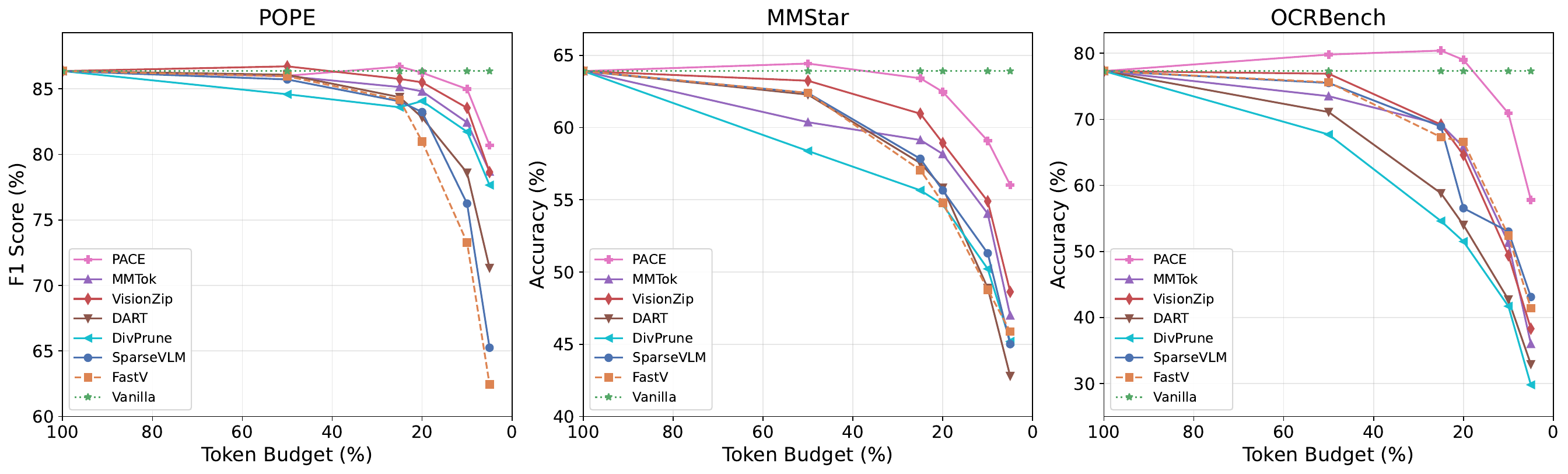}
 \caption{\textbf{Performance scaling under continuous token budgets.}
 PACE degrades more gracefully than pruning baselines as retention decreases, with especially large advantages on the detail-sensitive MMStar and OCRBench benchmarks under strict budgets.}
 \label{fig:token_scaling}
\end{figure*}

As summarized in Table~\ref{tab:main_results}, we report Qwen2.5-VL-7B performance under the strict fixed-resolution setting. Additional results for both the 3B and 7B variants are provided in Appendix~\ref{sec:appendix_add_exp}.

PACE consistently surpasses baselines across all configurations. Conventional post-encoder pruning methods suffer marked degradation on detail-sensitive tasks, notably DocVQA and OCRBench, particularly under aggressive 5\% or 10\% constraints. This vulnerability stems from the irreversible loss of high-frequency visual anchors---a direct consequence of relying predominantly on late-stage LLM relevance signals.

Conversely, the APC module condenses the visual representation while preserving text-dense and detail-sensitive regions prior to full encoding, enabling DDAE to precisely extract semantically relevant details. Under a highly restrictive 10\% token budget, PACE maintains 93.8\% of the full model's performance, yielding average performance gains of 12.7 and 13.9 percentage points over VisionZip and FastV, respectively. DocVQA still trails Vanilla because exact answers can depend on microscopic characters and layout relations that become ambiguous after condensation; nevertheless, PACE reaches 69.55, versus 59.56 for FastV and 49.55 for VisionZip. Small gains over Vanilla on a few tasks may result from removing distractors and are not our central claim.

Cross-backbone experiments in Appendix~\ref{sec:internvl_generalization} further evaluate InternVL3.5-4B's multi-tile pipeline. Across 25\%, 20\%, and 10\% retention, PACE attains normalized averages of 85.2\%, 80.9\%, and 69.5\%, exceeding the strongest corresponding baseline by 5.1, 4.7, and 4.0 points.

\paragraph{Robustness Validation.}
As illustrated in Figure~\ref{fig:token_scaling}, we evaluate the algorithmic robustness of PACE on Qwen2.5-VL-7B across continuous token budgets ranging from 100\% down to 5\%. We track performance on POPE, MMStar, and OCRBench, which assess object-level perception, visually grounded multimodal reasoning, and fine-grained OCR-centric recognition, respectively.

PACE exhibits robust performance scaling as the retention budget decreases. At the 10\% token budget, PACE achieves 84.99\% on POPE, 59.08\% on MMStar, and 70.90\% on OCRBench, outperforming the strongest baseline by 1.44, 4.18, and 17.90 absolute points. This advantage remains pronounced under the more aggressive 5\% budget, where PACE exceeds the best baseline by 2.00, 7.39, and 14.70 points across the three tasks. These results confirm that PACE successfully preserves both object-level perception and fine-grained visual evidence under severe compression.

\subsection{Efficiency Profiling}
\label{sec:efficiency}

We profile the inference efficiency of PACE on Qwen2.5-VL-7B under the fixed-resolution setting with a 10\% visual-token budget on a single RTX 4090. Since PACE does not accelerate autoregressive decoding, we report \textbf{time to first token (TTFT)} rather than end-to-end generation time. Vanilla TTFT comprises vision encoding and LLM prefill; PACE TTFT additionally includes the Shallow Feature Preview and adaptive resizing. The stage-wise encoder and prefill entries exclude these APC overheads.

PACE reduces the average vision encoder latency from 148.84 ms to 49.47 ms ($3.01\times$ speedup) and the LLM prefill latency from 217.05 ms to 32.69 ms ($6.64\times$ speedup). Although APC introduces an average 34.62 ms preview-and-resizing overhead, TTFT still decreases from 365.89 ms to 116.79 ms, a $3.13\times$ speedup. Across the full budget sweep in Appendix~\ref{sec:ttft_budget_sweep}, average TTFT speedup rises from $1.10\times$ at 80\% retention to $1.67\times$, $2.64\times$, and $3.13\times$ at 50\%, 20\%, and 10\%, respectively.

\begin{table}[ht]
\centering
\small
\setlength{\tabcolsep}{3.5pt}
\renewcommand{\arraystretch}{1.12}
\resizebox{\columnwidth}{!}{%
\begin{tabular}{l|cc|cc|cc}
\toprule
\multirow{2}{*}{\textbf{Dataset}} 
& \multicolumn{2}{c|}{\textbf{Encoder Time (ms)}} 
& \multicolumn{2}{c|}{\textbf{Prefill Time (ms)}} 
& \multicolumn{2}{c}{\textbf{TTFT incl. APC (ms)}} \\
& Vanilla / PACE & Spd. 
& Vanilla / PACE & Spd. 
& Vanilla / PACE & Spd. \\
\midrule
DocVQA & 143.41 / 50.74 & $2.83\times$ & 210.41 / 32.70 & $6.43\times$ & 353.82 / 117.54 & $3.01\times$ \\
TextVQA & 154.28 / 48.20 & $3.20\times$ & 223.68 / 32.69 & $6.84\times$ & 377.96 / 116.03 & $3.26\times$ \\
\midrule
\rowcolor[HTML]{E8F5E9}
\textbf{Avg.} & 148.84 / 49.47 & $\mathbf{3.01\times}$ & 217.05 / 32.69 & $\mathbf{6.64\times}$ & 365.89 / 116.79 & $\mathbf{3.13\times}$ \\
\bottomrule
\end{tabular}%
}
\vspace{1mm}
\caption{
\textbf{Latency of PACE on Qwen2.5-VL-7B at 10\% retention.}
 Encoder and prefill columns report their isolated stages; TTFT includes APC preview and resizing overhead. Measurements use the fixed-resolution setting on one RTX 4090.}
\label{tab:efficiency_analysis}
\end{table}

\section{Analysis and Discussion}
\label{sec:analysis_and_discussion}

In this section, we analyze the core mechanisms of PACE. Additional evaluations---including hyperparameter robustness and the shallow-feature preview ablation---are detailed in Appendix~\ref{sec:appendix_add_exp}.

\subsection{Orthogonality of APC}
\label{sec:apc_compatibility}

As a pre-encoder module, APC integrates orthogonally with existing post-encoder pruning algorithms. We validate this by augmenting VisionZip and MMTok with APC under 10\% and 5\% token-retention budgets. 

As shown in Table~\ref{tab:apc_compatibility}, APC improves both baselines, particularly on detail-sensitive benchmarks. Under a 10\% budget, APC boosts both VisionZip and MMTok by more than 15 points on ChartQA and OCRBench, alongside a gain of roughly 10 points on DocVQA. These improvements support APC's compatibility with different downstream extractors and its ability to preserve fine-grained visual evidence before encoding.

\begin{table}[ht]
\centering
\small
\renewcommand{\arraystretch}{1.15}
\resizebox{\columnwidth}{!}{%
\begin{tabular}{lc|cccccc}
\toprule
\textbf{Method} & \textbf{Budget} 
& \textbf{RealWorldQA } & \textbf{MME} & \textbf{ChartQA} & \textbf{OCRBench} & \textbf{TextVQA} & \textbf{DocVQA} \\
\midrule
VisionZip 
& 10\% 
& 65.23 & 2147 & 53.08 & 49.40 & 67.16 & 49.55 \\
\rowcolor[HTML]{E8F5E9}
\textbf{+ APC} 
& 10\% 
& \textbf{67.32} \up{2.09} 
& \textbf{2323} \up{176} 
& \textbf{68.96} \up{15.88} 
& \textbf{67.50} \up{18.10} 
& \textbf{72.39} \up{5.23} 
& \textbf{60.45} \up{10.90} \\
\midrule
MMTok 
& 10\% 
& 58.82 & 2218 & 51.04 & 51.30 & 67.33 & 51.37 \\
\rowcolor[HTML]{E8F5E9}
\textbf{+ APC} 
& 10\% 
& \textbf{63.27} \up{4.45} 
& \textbf{2289} \up{71} 
& \textbf{66.84} \up{15.80} 
& \textbf{66.80} \up{15.50} 
& \textbf{72.56} \up{5.23} 
& \textbf{62.05} \up{10.68} \\
\midrule
VisionZip 
& 5\% 
& 61.57 & 2029 & 40.96 & 38.30 & 55.47 & 29.40 \\
\rowcolor[HTML]{E8F5E9}
\textbf{+ APC} 
& 5\% 
& \textbf{64.18} \up{2.61} 
& \textbf{2206} \up{177} 
& \textbf{48.88} \up{7.92} 
& \textbf{50.10} \up{11.80} 
& \textbf{58.79} \up{3.32} 
& \textbf{37.44} \up{8.04} \\
\midrule
MMTok 
& 5\% 
& 54.12 & 2133 & 32.76 & 36.00 & 50.95 & 28.85 \\
\rowcolor[HTML]{E8F5E9}
\textbf{+ APC} 
& 5\% 
& \textbf{59.48} \up{5.36} 
& \textbf{2221} \up{88} 
& \textbf{48.56} \up{15.80} 
& \textbf{50.70} \up{14.70} 
& \textbf{60.40} \up{9.45} 
& \textbf{40.64} \up{11.79} \\
\bottomrule
\end{tabular}%
}
\vspace{1mm}
\caption{
\textbf{Orthogonal integration of APC.}
 APC augments VisionZip and MMTok at 10\% and 5\% token budgets.}
\label{tab:apc_compatibility}
\end{table}

\begin{figure*}[t]
    \centering
    \includegraphics[width=0.88\textwidth]{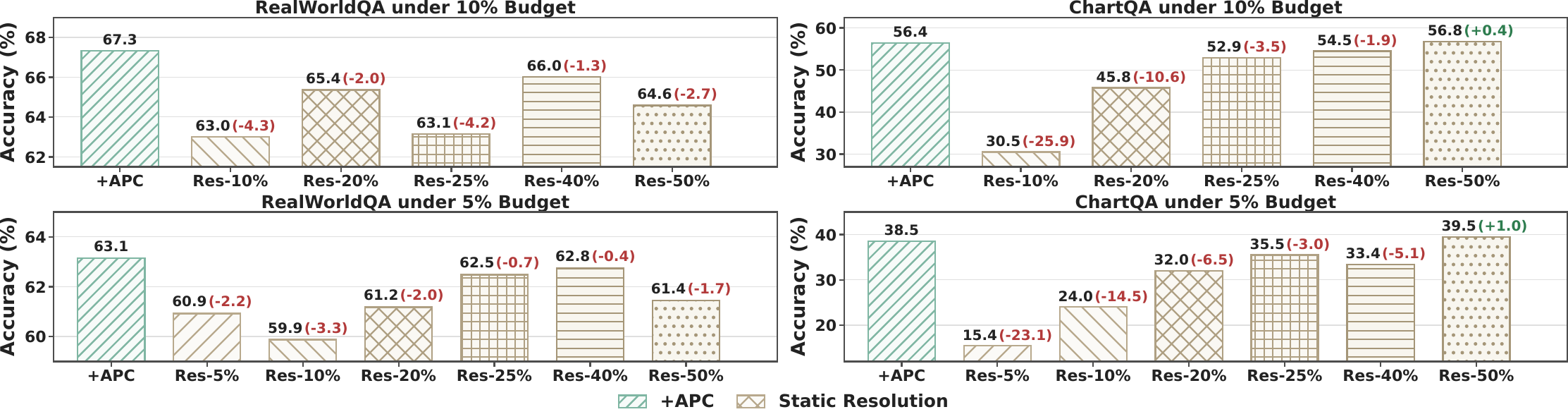}
    \caption{
    \textbf{Adaptive vs. static resolution under the dynamic-resolution setting.}
    Adaptive pixel allocation is compared with static-resolution baselines at 10\% and 5\% token budgets.}
    \label{fig:adaptive_vs_fixed_dynamic}
\end{figure*}

\subsection{Adaptive Resolution vs. Static Resolution}
\label{sec:adaptive_vs_fixed_res_dynamic}

We evaluate the efficacy of APC's adaptive pixel allocation against static-resolution variants under the dynamic-resolution setting of Qwen2.5-VL-7B. While static-resolution methods apply rigid condensation ratios globally, APC dynamically modulates the input resolution based on global density and local detail metrics.

As illustrated in Figure~\ref{fig:adaptive_vs_fixed_dynamic}, Adaptive Resolution consistently achieves a favorable performance balance across diverse domains. On holistic tasks such as RealWorldQA, it outperforms the best static-resolution counterpart at both 10\% and 5\% budgets. Conversely, on detail-sensitive tasks such as ChartQA, Fixed Res.-50 yields a marginal gain over adaptive resolution (0.36 points at 10\%) but concurrently induces a significant drop (2.74 points) on RealWorldQA. This trade-off confirms that rigid global scaling inherently overfits specific domains at the expense of general visual robustness, whereas APC successfully balances broad layout contexts with microscopic visual evidence.

\subsection{Effect of Attention Fusion in DDAE}
\label{sec:ddae_attention_fusion}

We evaluate DDAE's dynamic fusion against single-modality and fixed-weight baselines: \textbf{DDAE}, \textbf{Only ViT-Attn} (solely vision-side visual attention), \textbf{Only LLM-Attn} (solely LLM-side semantic attention), and \textbf{Fixed Fusion} (a static combination that assigns a weight of 0.5 to each signal).

\begin{figure}[ht]
 \centering
 \includegraphics[width=0.92\columnwidth]{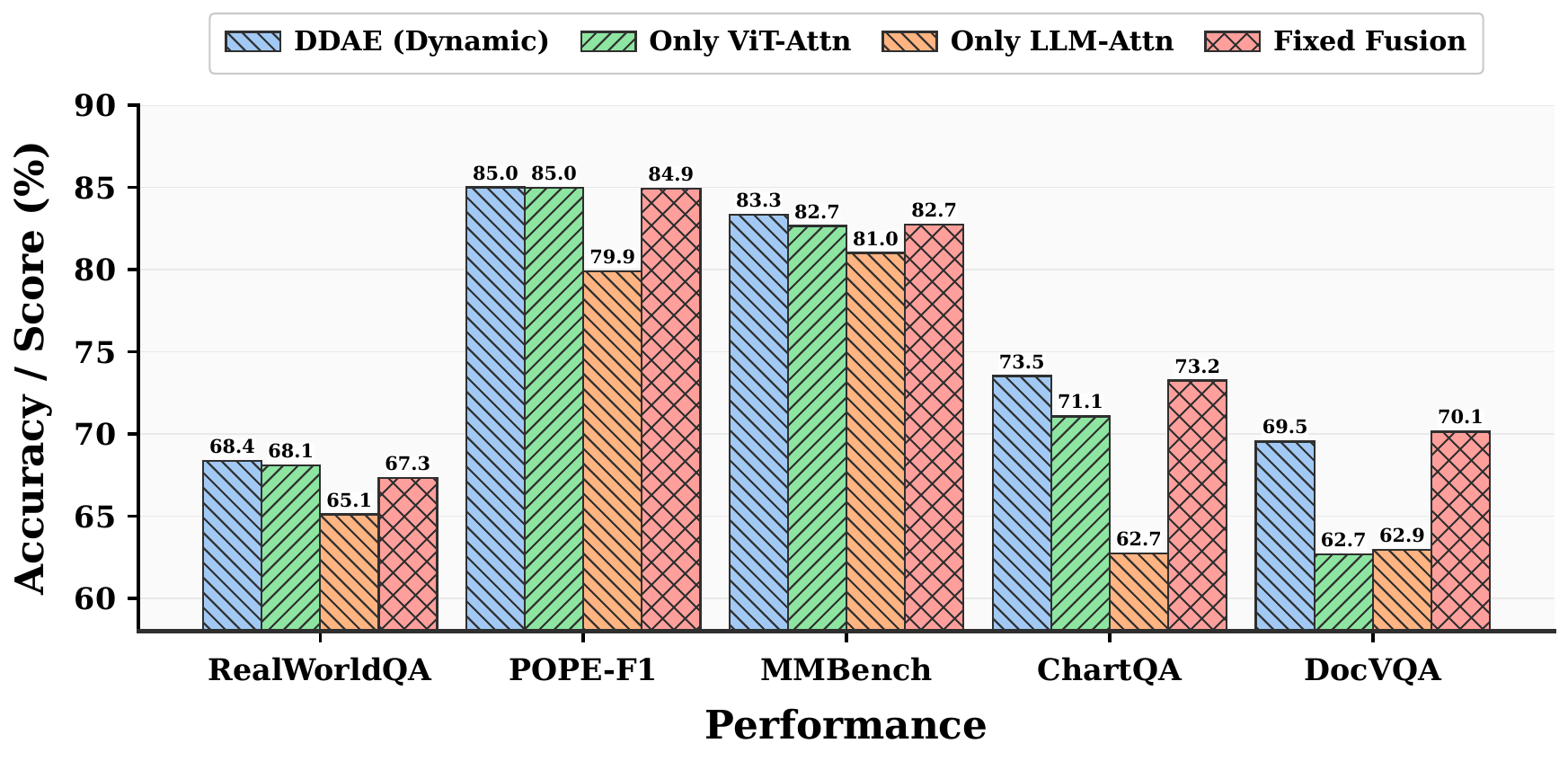}
 \caption{
 \textbf{Effect of attention fusion in DDAE.}
 Dynamic fusion yields the most balanced performance across benchmarks: language-only attention loses critical visual evidence, while vision-only attention weakens task alignment.}
 \label{fig:ddae_attention_fusion}
\end{figure}

As shown in Figure~\ref{fig:ddae_attention_fusion}, Dynamic DDAE exhibits the most resilient performance profile. Relying exclusively on LLM attention incurs substantial drops on visual tasks, with ChartQA degrading by over 10 points, indicating that text-guided signals neglect unprompted yet vital visual contexts. Conversely, omitting text-guided semantics (Only ViT-Attn) compromises task alignment. These representative gaps affirm that robust feature extraction necessitates dynamic, confidence-weighted multimodal attention. 

\subsection{Impact of DDAE Extraction Depth}
\label{sec:impact_ddae_depth}

The LLM extraction depth ($L_{ext}$) governs the trade-off between prefill latency and visual reasoning. A shallower layer initiates earlier pruning, significantly reducing prefill latency, whereas a deeper layer yields more sophisticated cross-modal attention maps at the cost of processing the full visual sequence longer. Table~\ref{tab:ddae_depth_ablation} illustrates this trade-off under a 10\% token budget.

Deferring DDAE extraction to deeper layers (e.g., depth 24) enhances visual reasoning, markedly boosting ChartQA performance over depth 8. However, under the $2048\times28\times28$ fixed-resolution setting, the prefill speedup falls from $2.11\times$ at layer 2 to only $1.09\times$ at layer 24. The same trend holds at twice the resolution (Appendix~\ref{sec:depth_latency_full}). We therefore use layer 2 ($L_{ext}=2$) as the efficiency-oriented default.

\begin{table}[ht]
\centering
\small
\renewcommand{\arraystretch}{1.15}
\resizebox{\columnwidth}{!}{%
\begin{tabular}{c|cccc|cc}
\toprule
\textbf{Depth ($L_{ext}$)} 
& \textbf{RealWorldQA} 
& \textbf{POPE} 
& \textbf{ChartQA} 
& \textbf{TextVQA}
& \textbf{Prefill (ms)}
& \textbf{Spd.} \\
\midrule
\rowcolor[HTML]{E8F5E9}
\textbf{2}  
& \textbf{68.37} & 84.99 & 73.52 & 78.42 & \textbf{35.26} & $\mathbf{2.11\times}$ \\
8  
& 66.54 & 84.72 & 65.68 & 78.80 & 42.51 & $1.75\times$ \\
16 
& 66.93 & 85.31 & 70.60 & \textbf{79.24} & 55.32 & $1.35\times$ \\
24 
& 67.71 & \textbf{86.32} & \textbf{78.96} & 78.41 & 68.12 & $1.09\times$ \\
\bottomrule
\end{tabular}%
}
\vspace{1mm}
\caption{
\textbf{Quality--latency trade-off of DDAE extraction depth.}
 Prefill latency is compared against the 74.52 ms PACE-without-DDAE baseline.}
\label{tab:ddae_depth_ablation}
\end{table}

\subsection{Impact of Attention Token Sources}
\label{sec:impact_attention_tokens}

We compare three semantic cross-attention aggregation strategies for DDAE: \textbf{Vision + Text} (aggregating over all sequence tokens), \textbf{Text} (restricting strictly to linguistic tokens), and \textbf{Last} (relying exclusively on the terminal token).

\begin{figure}[ht]
 \centering
 \includegraphics[width=0.82\columnwidth]{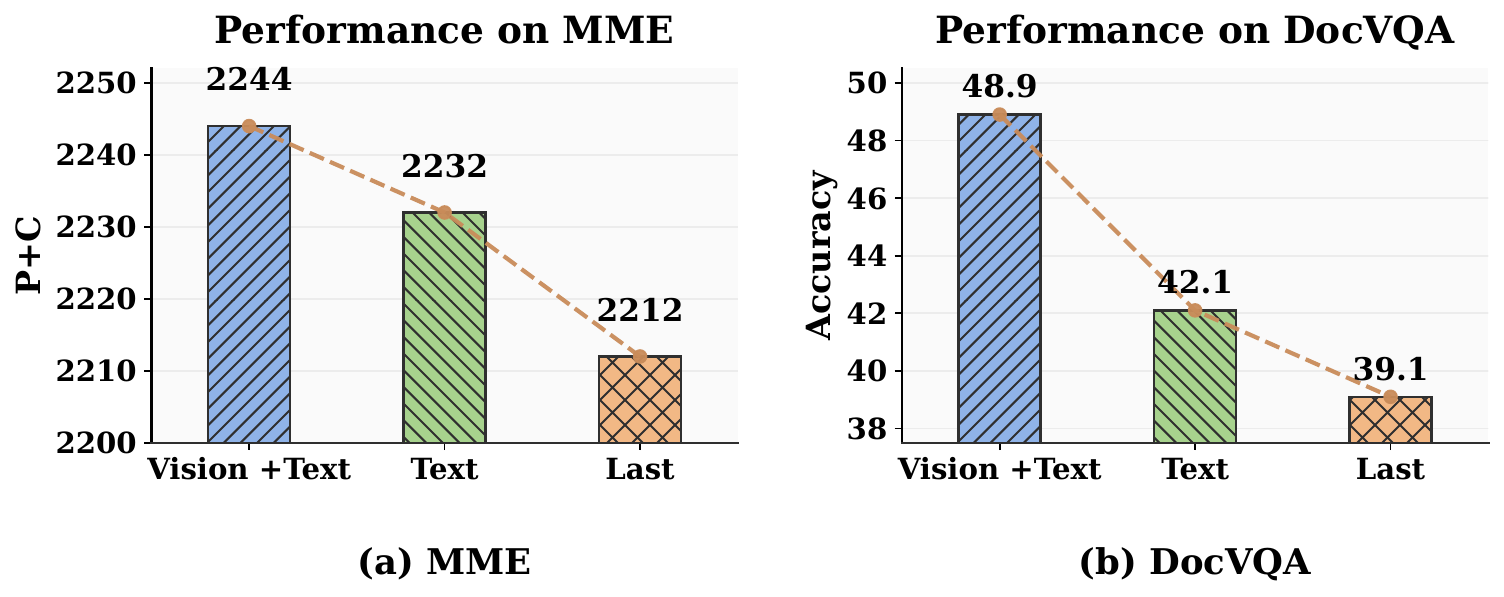}
 \caption{\textbf{Attention-token sources for DDAE.}
We compare semantic attention aggregated from the complete sequence, text tokens only, or the final token. Complete aggregation best preserves both MME perception and DocVQA layout evidence.}
 \label{fig:attention_token_source}
\end{figure}

Figure~\ref{fig:attention_token_source} shows that the \textbf{Vision + Text} strategy performs best on both MME and DocVQA. Restricting aggregation to linguistic or terminal tokens progressively reduces performance, with DocVQA dropping by nearly 10 points. This result suggests that complete-sequence aggregation better preserves the spatial and visual evidence needed for layout-sensitive comprehension.

\section{Conclusion}
This paper introduces \textbf{PACE}, a training-free inference framework designed to accelerate high-resolution VLMs via a unified \textbf{Condense-and-Extract} paradigm. To overcome the dual computational bottlenecks of the vision encoder and the LLM, the \textit{Condense} phase uses APC to adaptively remove pixel-level redundancy prior to visual encoding, mitigating compute-bound ViT overhead while preserving global layouts. Subsequently, the \textit{Extract} phase employs DDAE to retain salient fine-grained details by dynamically fusing internal visual priors from the ViT with semantic relevance from the LLM. PACE retains 93.8\% of Qwen2.5-VL-7B's uncompressed performance at a 90\% token reduction and delivers a $3.1\times$ TTFT speedup.

\enlargethispage{2\baselineskip}
\section{Limitations}
\label{sec:limitations}

PACE has two important limitations. First, APC accelerates the encoder only when reducing the pixel or tile budget also reduces the number of encoder tokens. On fixed-grid VLMs, DDAE can still reduce LLM prefill cost, but APC provides no encoder-side gain. Preview overhead also varies with architecture and resolution, so the Qwen2.5-VL speedups may not transfer unchanged to other backbones.

Second, APC performs query-agnostic, one-shot condensation and can miss faint or tiny characters, small chart labels, thin lines, or low-contrast objects. Once resizing makes such evidence ambiguous, DDAE cannot reconstruct it. High-stakes applications should therefore use a higher retention floor, a less aggressive target budget, or a lower $\alpha$ to emphasize local detail. Uncertainty-triggered or query-conditioned recovery of high-resolution crops is a promising extension.

\clearpage
\bibliography{custom} 

\clearpage
\appendix

\section{Detailed Experimental Setup}
\label{sec:appendix_setup}

\subsection{Models, Baselines, and Evaluation Framework}

To ensure fair and reproducible comparisons, we detail the backbone models, token-reduction baselines, and evaluation framework used in our experiments.

\paragraph{Models.}
We integrate PACE into \textbf{Qwen2.5-VL-3B}\footnote{\url{https://huggingface.co/Qwen/Qwen2.5-VL-3B-Instruct}} and \textbf{Qwen2.5-VL-7B}\footnote{\url{https://huggingface.co/Qwen/Qwen2.5-VL-7B-Instruct}} \citep{bai2025qwen25vltechnicalreport}. Qwen2.5-VL uses native dynamic-resolution processing to convert images of different sizes into variable-length visual token sequences. Its visual backbone combines a dynamic-resolution Vision Transformer with Window Attention to reduce self-attention cost. Because visual-token length still grows with input resolution, high-resolution inputs remain expensive to encode. This design makes Qwen2.5-VL suitable for evaluating PACE's adaptive pixel allocation before full visual encoding.

\paragraph{Baselines.}
We compare PACE with six visual-token reduction methods:
\begin{itemize}
 \item \textbf{FastV} \citep{chen2024image}: Discards visual tokens exhibiting minimal attention scores within early LLM layers, thereby truncating the context sequence for subsequent layers.
 \item \textbf{SparseVLM} \citep{zhang2024sparsevlm}: Deploys text-guided, training-free sparsification utilizing decoder self-attention to assess token importance, paired with a token recycling mechanism.
 \item \textbf{DivPrune} \citep{alvar2025divprune}: Formulates token pruning mathematically as a Max-Min Diversity Problem, ensuring the selected subset remains visually and semantically heterogeneous.
 \item \textbf{DART} \citep{wen2025stop}: Evaluates contextual duplication to purge visual tokens highly redundant with selected pivot tokens while preserving distinct visual representations.
 \item \textbf{VisionZip} \citep{yang2025visionzip}: Attenuates sequences through sequential token selection and fusion, merging redundant local patches into compressed contextual representations.
 \item \textbf{MMTok} \citep{dong2025mmtok}: Models visual-token selection explicitly as a submodular maximum coverage problem, employing greedy heuristics to optimize conceptual graph overlap.
\end{itemize}

All six baselines operate after visual representation extraction. They reduce LLM prefill cost but do not change the cost of encoding the original high-resolution input with the visual backbone.

\paragraph{Evaluation Framework.}
We standardize all experiments using the open-source \textbf{\texttt{lmms-eval}} framework \citep{zhang2025lmms}. This pipeline provides a unified evaluation suite encompassing standardized task definitions, exact prompt templates, deterministic generation configurations, and uniform metric computation. Unless explicitly stated otherwise, we adhere strictly to the official \texttt{lmms-eval} configuration to ensure direct comparability across all methods.

\subsection{Detailed Benchmark Descriptions}

We comprehensively evaluate PACE across nine diverse benchmarks covering disparate visual processing demands, with their statistical properties cataloged in Table~\ref{tab:dataset_full_stats}.
\begin{itemize}
 \item \textbf{RealWorldQA} \citep{grok15}: Evaluates general visual understanding in real-world scenes, including object localization and scene-level awareness.
 \item \textbf{POPE} \citep{li2023evaluating}: Measures object hallucination under different visual conditions.
 \item \textbf{MME} \citep{fu2026mme}: Covers perception, OCR, logical reasoning, and commonsense tasks.
 \item \textbf{MMBench} \citep{liu2024mmbench}: Uses multiple-choice questions to evaluate a range of multimodal capabilities.
 \item \textbf{MMStar} \citep{chen2024we}: Contains 1,500 curated questions designed to reduce language-only shortcuts and require visual evidence.
 \item \textbf{OCRBench} \citep{liu2024ocrbench}: Evaluates OCR-related capabilities across scene text, documents, and mathematical content.
 \item \textbf{TextVQA} \citep{singh2019towards}: Evaluates reading and reasoning over text embedded in natural images.
 \item \textbf{ChartQA} \citep{masry2022chartqa}: Evaluates numerical and structural reasoning over charts.
 \item \textbf{DocVQA} \citep{mathew2021docvqa}: Evaluates document reading and layout understanding on scanned documents.
\end{itemize}

\begin{table}[htbp]
\centering
\resizebox{\columnwidth}{!}{%
\begin{tabular}{l r c r r r}
\toprule
\textbf{Dataset} & \textbf{Total} & \textbf{Avg. Resolution} &
\textbf{Avg. Pixels} & \textbf{Duplicates} & \textbf{Dup. Ratio} \\
\midrule
RealWorldQA & 765 & $1316 \times 1030$ & 1,341,566 & 3 & 0.39\% \\
OCRBench & 1,000 & $615 \times 732$ & 1,017,574 & 70 & 7.00\% \\
MMStar & 1,500 & $511 \times 391$ & 245,161 & 70 & 4.67\% \\
MME & 2,374 & $1086 \times 945$ & 1,881,888 & 1,197 & 50.42\% \\
ChartQA & 2,500 & $768 \times 583$ & 454,992 & 991 & 39.64\% \\
MMBench$_{\mathrm{en\_dev}}$
 & 4,329 & $440 \times 338$ & 153,326 & 3,208 & 74.10\% \\
TextVQA$_{\mathrm{val}}$
 & 5,000 & $952 \times 819$ & 770,513 & 1,834 & 36.68\% \\
DocVQA$_{\mathrm{val}}$
 & 5,349 & $1783 \times 2099$ & 3,935,008 & 4,064 & 75.98\% \\
POPE & 9,000 & $585 \times 479$ & 277,243 & 8,500 & 94.44\% \\
\bottomrule
\end{tabular}%
}
\caption{\textbf{Statistics of the evaluation datasets.}
The table reports dataset size, native resolution, pixel count, and duplication statistics. ``Duplicates'' counts additional question instances that reuse an image already paired with another question.}
\label{tab:dataset_full_stats}
\end{table}

\subsection{Detailed Implementation Specifics}

For the APC module, we fix the feature preview depth to the initial ViT block ($K=1$) to limit preprocessing overhead. We set the global-local balancing weight $\alpha$ to $0.6$ and the local contrast regularization term $\gamma$ to $1.5$. The target retention ratio $r = \rho$ determines the final pixel count relative to the original image area. Consequently, the image width and height are both scaled by a factor of $\sqrt{r}$ using bicubic interpolation (\texttt{Image.Resampling.BICUBIC}).

In the DDAE module, the visual self-attention score ($S_{vis}$) comes directly from the final ViT block. The semantic relevance map ($S_{llm}$) is aggregated from the cross-modal attention of the second LLM block ($L_{ext}=2$), with weights averaged uniformly across all valid tokens. We set the temperature scalar $\tau$ for confidence-weighted softmax fusion to $0.5$. All experiments use greedy decoding (temperature $=0$). We conduct all efficiency profiling and latency measurements, including TTFT and TPOT calculations, on a single NVIDIA RTX 4090 GPU.

\section{Additional Experiments}
\label{sec:appendix_add_exp}

\subsection{Hyperparameter Robustness}
\label{sec:hyperparameter_robustness}

PACE uses a small set of interpretable hyperparameters. The scaling variable $\gamma$ limits over-retention caused by noisy, high-frequency patterns in the local contrast score. The temperature $\tau$ controls the sharpness of DDAE's softmax confidence weights.

\paragraph{Sensitivity of Global-Local Balancing Weight ($\alpha$).}
\label{sec:alpha_sensitivity}

The coefficient $\alpha$ controls the balance between global redundancy estimation and local detail preservation in APC. Specifically, a larger $\alpha$ places greater emphasis on the global information density score $\rho_g$, whereas a smaller $\alpha$ increases the contribution of the local detail contrast score $\rho_d$. To examine this trade-off, we evaluate different $\alpha$ values under the restrictive 5\% token budget in the dynamic-resolution setting of Qwen2.5-VL-7B.

Table~\ref{tab:mmstar_alpha_sensitivity} reports the fine-grained MMStar breakdown. Although larger values such as $\alpha=1.0$ and $\alpha=0.8$ improve specific categories, including coarse perception and math reasoning, they consistently degrade several layout- and relation-sensitive dimensions. For example, compared with the default $\alpha=0.6$, setting $\alpha=1.0$ decreases instance reasoning by 3.61 points, logical reasoning by 4.35 points, and science \& technology by 3.03 points.

Conversely, smaller values such as $\alpha=0.4$ and $\alpha=0.2$ over-emphasize local contrast cues. This makes the adaptive resolution policy more sensitive to isolated high-frequency patterns and weakens its ability to preserve coherent global layouts. As a result, these settings reduce the MMStar average by 3.32 and 2.90 points, respectively. Overall, $\alpha=0.6$ provides the most balanced behavior, securing a stable compromise between suppressing globally redundant regions and retaining task-critical local cues under severe token constraints.

\begin{table*}[ht]
\centering
\small
\renewcommand{\arraystretch}{1.15}
\resizebox{\textwidth}{!}{%
\begin{tabular}{c|ccccccc}
\toprule
\textbf{$\alpha$} 
& \textbf{Coarse Perc.} 
& \textbf{Fine-grained Perc.} 
& \textbf{Instance Reason.} 
& \textbf{Logical Reason.} 
& \textbf{Math} 
& \textbf{Science \& Tech.} 
& \textbf{Avg.} \\
\midrule
\rowcolor[HTML]{E8F5E9}
\textbf{0.6 (Default)} 
 & 59.48 \eq
& \textbf{39.07} \eq
& \textbf{57.63} \eq
& \textbf{48.47} \eq
& 43.45 \eq
& \textbf{33.06} \eq
& \textbf{46.86} \eq \\

1.0 
 & \textbf{62.00} \up{2.52}
& 37.25 \dn{1.82}
& 54.02 \dn{3.61}
& 44.12 \dn{4.35}
& 46.25 \up{2.80}
& 30.03 \dn{3.03}
& 45.16 \dn{1.70} \\

0.8 
& 60.35 \up{0.87}
& 33.65 \dn{5.42}
& 53.64 \dn{3.99}
& 45.63 \dn{2.84}
& \textbf{50.02} \up{6.57}
& 27.52 \dn{5.54}
& 45.14 \dn{1.72} \\

0.4 
& 58.32 \dn{1.16}
& 36.42 \dn{2.65}
& 53.83 \dn{3.80}
& 41.77 \dn{6.70}
& 44.24 \up{0.79}
& 26.68 \dn{6.38}
& 43.54 \dn{3.32} \\

0.2 
& 58.52 \dn{0.96}
& 33.41 \dn{5.66}
& 52.57 \dn{5.06}
& 44.56 \dn{3.91}
& 43.84 \up{0.39}
& 30.84 \dn{2.22}
& 43.96 \dn{2.90} \\
\bottomrule
\end{tabular}%
}
\vspace{1mm}
\caption{
\textbf{Sensitivity of APC's global-local balancing weight $\alpha$ on MMStar under the dynamic-resolution setting.}
Results are reported on Qwen2.5-VL-7B at the 5\% token budget.}
\label{tab:mmstar_alpha_sensitivity}
\end{table*}

\paragraph{Sensitivity of the Top-Detail Percentile.}
\label{sec:tail_percentile}
We compare 5\%, 10\%, and 20\% tail fractions while holding all other settings fixed. Their performance is very similar, so we use 10\% as a fixed midpoint across models, datasets, and token budgets.

\subsection{Shallow Feature Preview Ablation}
\label{sec:preview_ablation}

We isolate the preview signal while keeping APC's adaptive resizing and DDAE unchanged under the native dynamic-resolution setting at 10\% visual-token retention. A literal removal of all preview signals would make image-dependent allocation impossible and reduce APC to static resolution; we therefore replace the shallow ViT feature with four training-free pixel statistics.

\begin{table}[ht]
\centering
\small
\setlength{\tabcolsep}{5pt}
\renewcommand{\arraystretch}{1.12}
\begin{tabular}{lcc}
\toprule
\textbf{APC Feature} & \textbf{RealWorldQA} & \textbf{ChartQA} \\
\midrule
Vanilla (100\%) & 69.54 & 83.92 \\
RGB statistics & 63.40 & 47.64 \\
Color entropy & 65.10 & 51.40 \\
Edge density & 64.18 & 55.96 \\
Laplacian & 64.31 & 55.04 \\
\rowcolor[HTML]{E8F5E9}
\textbf{Shallow Feature Preview} & \textbf{67.32} & \textbf{56.40} \\
\bottomrule
\end{tabular}
\caption{\textbf{Ablation of APC's information-density feature.}
 Compressed variants retain 10\% of visual tokens on Qwen2.5-VL-7B; Vanilla is shown only as the uncompressed reference. The shallow ViT preview is the strongest adaptive feature on both a natural-image and a chart benchmark.}
\label{tab:preview_ablation}
\end{table}

Pixel statistics measure color variation or local high-frequency response but cannot reliably separate semantic information density from texture, noise, decorative patterns, or irrelevant edges. Edge density is comparatively effective on ChartQA because plotted lines convey useful structure, whereas the Laplacian is more sensitive to fine-scale noise. The shallow ViT preview provides a lightweight semantic prior and transfers more consistently across the two domains.

\subsection{Cross-Model Generalization to InternVL3.5-4B}
\label{sec:internvl_generalization}

We evaluate PACE on InternVL3.5-4B, whose dynamic multi-tile pipeline differs substantially from Qwen-style native-resolution patching. For applicable baselines, ``-T'' prunes within each tile before concatenation, whereas ``-G'' concatenates all tile tokens before one global selection under the same total budget. We report both reasonable treatments of tile boundaries.

\begin{table*}[htbp]
\centering
\small
\setlength{\tabcolsep}{2.5pt}
\renewcommand{\arraystretch}{1.08}
\resizebox{\textwidth}{!}{%
\begin{tabular}{l|ccccccccc|c}
\toprule
\textbf{Method} & \textbf{RWQA} & \textbf{POPE} & \textbf{MME} & \textbf{MMB} & \textbf{MMStar} & \textbf{ChartQA} & \textbf{OCRB} & \textbf{TextVQA} & \textbf{DocVQA} & \textbf{Avg.} \\
\midrule
\rowcolor[HTML]{FAFAFA}
Vanilla (100\%) & 65.62 & 89.47 & 2280.88 & 81.36 & 66.53 & 85.72 & 80.80 & 75.82 & 91.08 & 100.0\% \\
\midrule
\multicolumn{11}{c}{\textit{Retain 25\% visual tokens}} \\
FastV & 57.91 & 87.24 & 2108.89 & 76.46 & 54.09 & \textbf{59.68} & 38.10 & 66.39 & 57.65 & 80.1\% \\
DivPrune-T & 58.04 & 87.97 & 2011.20 & 74.74 & 52.34 & 48.16 & 30.30 & 56.58 & 42.05 & 73.3\% \\
DivPrune-G & 56.86 & 87.65 & 2026.90 & 77.58 & 54.81 & 50.36 & 34.00 & 57.05 & 46.54 & 75.4\% \\
VisionZip-T & \textbf{58.43} & 88.25 & 1922.14 & 73.71 & 50.69 & 31.72 & 16.10 & 45.22 & 24.05 & 64.6\% \\
MMTok-T & 57.65 & 88.17 & 2023.92 & 74.66 & 51.08 & 38.72 & 12.10 & 34.89 & 27.26 & 64.4\% \\
MMTok-G & 58.04 & \textbf{88.42} & 2049.51 & 76.46 & 51.82 & 40.16 & 12.70 & 31.89 & 29.59 & 65.1\% \\
\rowcolor[HTML]{E8F5E9}
\textbf{PACE (Ours)} & 58.17 & 88.03 & \textbf{2238.93} & \textbf{80.58} & \textbf{58.47} & 55.64 & \textbf{52.30} & \textbf{66.46} & \textbf{70.36} & \textbf{85.2\%} \\
\midrule
\multicolumn{11}{c}{\textit{Retain 20\% visual tokens}} \\
FastV & 56.21 & 85.95 & 2032.87 & 74.66 & 52.17 & \textbf{51.88} & 33.70 & \textbf{63.91} & 52.76 & 76.2\% \\
DivPrune-T & 55.29 & 87.45 & 2002.21 & 73.80 & 50.08 & 42.48 & 23.30 & 53.15 & 36.80 & 69.4\% \\
DivPrune-G & 54.12 & 86.79 & 1966.18 & 76.12 & 52.39 & 44.68 & 29.70 & 53.15 & 40.31 & 71.2\% \\
VisionZip-T & 55.82 & \textbf{88.25} & 1884.68 & 73.37 & 49.56 & 26.80 & 11.70 & 39.89 & 19.31 & 61.2\% \\
MMTok-T & 55.82 & 87.57 & 1967.05 & 72.42 & 48.57 & 32.88 & 8.50 & 31.44 & 23.33 & 60.8\% \\
MMTok-G & \textbf{56.60} & 87.78 & 1968.71 & 74.48 & 48.43 & 32.40 & 9.80 & 28.94 & 24.49 & 61.1\% \\
\rowcolor[HTML]{E8F5E9}
\textbf{PACE (Ours)} & 55.42 & 87.55 & \textbf{2183.78} & \textbf{78.44} & \textbf{57.30} & 48.24 & \textbf{45.80} & 63.43 & \textbf{64.62} & \textbf{80.9\%} \\
\midrule
\multicolumn{11}{c}{\textit{Retain 10\% visual tokens}} \\
FastV & 51.11 & 83.33 & 1804.05 & 71.13 & 43.06 & \textbf{36.80} & 21.80 & \textbf{56.39} & 39.00 & 65.5\% \\
DivPrune-T & 51.50 & 86.19 & 1853.69 & 70.79 & 45.53 & 30.00 & 14.00 & 43.09 & 25.23 & 60.9\% \\
DivPrune-G & 51.63 & 85.16 & 1851.55 & 71.48 & 46.99 & 29.36 & 17.50 & 43.77 & 28.13 & 62.0\% \\
VisionZip-T & 50.46 & 86.33 & 1819.59 & 67.70 & 43.35 & 20.20 & 7.60 & 26.03 & 13.99 & 53.8\% \\
MMTok-T & 53.20 & 86.08 & 1848.30 & 69.07 & 43.23 & 23.00 & 6.10 & 25.41 & 16.06 & 54.9\% \\
MMTok-G & \textbf{54.38} & \textbf{86.42} & 1924.16 & 70.53 & 44.06 & 21.92 & 6.00 & 22.39 & 17.76 & 55.4\% \\
\rowcolor[HTML]{E8F5E9}
\textbf{PACE (Ours)} & 52.16 & 84.55 & \textbf{2083.74} & \textbf{75.34} & \textbf{51.35} & 30.04 & \textbf{29.20} & 53.86 & \textbf{43.82} & \textbf{69.5\%} \\
\bottomrule
\end{tabular}%
}
\caption{\textbf{Cross-model performance on InternVL3.5-4B.}
 Benchmark scores are reported for a variable-token, multi-tile architecture at 25\%, 20\%, and 10\% retention. Avg. first normalizes each score by the corresponding Vanilla score and then averages the nine ratios; bold marks the best compressed result within each budget.}
\label{tab:internvl_results}
\end{table*}

PACE achieves the strongest overall performance, surpassing the best baseline average by 5.1, 4.7, and 4.0 points at 25\%, 20\%, and 10\%, respectively. It is not uniformly best on every task, but it provides the most balanced result and preserves substantially more OCRBench and DocVQA performance. These results support transfer to a distinct tiling-based architecture; latency and preview amortization remain backbone- and resolution-dependent.

\subsection{Comprehensive Experimental Results}

We report complete results for \textbf{Qwen2.5-VL-7B} (Table~\ref{tab:appendix_7b}) and \textbf{Qwen2.5-VL-3B} (Table~\ref{tab:appendix_3b}) under two input settings:
\begin{itemize}
 \item \textbf{Fixed-resolution setting (MinPix = MaxPix = $2048 \times 28 \times 28$):} Pads or resizes every input to the same pixel budget, which standardizes the input size and can introduce additional redundant patches.
 \item \textbf{Dynamic-resolution setting (MinPix = $256 \times 28 \times 28$, MaxPix = $2048 \times 28 \times 28$):} Uses the model's native resolution range while preserving each image's aspect ratio.
\end{itemize}

Across both settings, post-encoder baselines degrade substantially under the \textbf{fixed-resolution setting} at 10\% and 5\% retention, particularly on detail-sensitive benchmarks. PACE degrades more gradually across both model sizes, consistent with the benefit of adapting the input resolution before full visual encoding.

\begin{table*}[htbp]
\centering
\small
\renewcommand{\arraystretch}{1.1}
\resizebox{\textwidth}{!}{%
\begin{tabular}{l|ccccccccc|c}
\toprule
\textbf{Method} & \textbf{RealWorldQA} & \textbf{POPE} & \textbf{MME} & \textbf{MMBench} & \textbf{MMStar} & \textbf{ChartQA}& \textbf{OCRBench}& \textbf{TextVQA} & \textbf{DocVQA} & \textbf{Avg.} \\
& Acc. $\uparrow$ & F1 $\uparrow$ & P+C $\uparrow$ & Acc. $\uparrow$ & Acc. $\uparrow$ & Acc. $\uparrow$ & Acc. $\uparrow$ & Acc. $\uparrow$ & ANLS $\uparrow$ & $\uparrow$ \\
\midrule
\rowcolor[HTML]{E8EAF6} \multicolumn{11}{c}{\textbf{Fixed-resolution setting (MinPix = $2048 \times 28 \times 28$, MaxPix = $2048 \times 28 \times 28$)}} \\
\midrule
\rowcolor[HTML]{EFEFEF} \textit{Vanilla (100\% Tokens)} & 69.54 & 86.36 & 2317 & 82.99 & 63.91 & 78.20 & 77.30 & 82.37 & 94.74 & 100.0\% \\
\midrule
\rowcolor[HTML]{FAFAFA} \multicolumn{11}{c}{\textit{Retain 20\% $\bar{T}$}} \\
FastV (ECCV'24) & 64.58 & 80.99 & 2256 & 80.76 & 54.79 & 64.12 & 66.60 & 79.16 & 76.58 & 90.2\% \\
SparseVLM (ICML'25) & 66.41 & 83.23 & 2258 & 81.36 & 55.65 & 70.00 & 56.54 & 80.33 & 73.56 & 90.3\% \\
DivPrune (CVPR'25) & 61.83 & 84.07 & 2248 & 80.07 & 54.66 & 51.56 & 51.50 & 69.99 & 49.40 & 81.7\% \\
DART (EMNLP'25) & 63.53 & 82.81 & 2278 & 79.64 & 55.83 & 58.72 & 54.00 & 69.48 & 50.58 & 83.5\% \\
VisionZip (CVPR'25) & \textbf{67.06} & 85.50 & 2317 & 81.01 & 58.93 & 69.68 & 64.60 & 77.52 & 75.77 & 92.5\% \\
MMTok (ICLR'26) & 63.79 & 84.81 & 2278 & 82.22 & 58.17 & 68.40 & 65.80 & 76.78 & 74.25 & 91.4\% \\
PACE (w/o APC) & 66.93 & 85.29 & 2310 & 81.36 & 59.29 & 72.84 & 69.10 & 80.72 & 83.00 & 94.9\% \\
\rowcolor[HTML]{E8F5E9} \textbf{PACE (Ours)} & 66.93 & \textbf{86.23} & \textbf{2322} & \textbf{83.08} & \textbf{62.47} & \textbf{78.56} & \textbf{79.00} & \textbf{81.66} & \textbf{86.84} & \textbf{98.6\%} \\
\midrule
\rowcolor[HTML]{FAFAFA} \multicolumn{11}{c}{\textit{Retain 10\% $\bar{T}$}} \\
FastV (ECCV'24) & 59.08 & 73.28 & 2154 & 77.23 & 48.78 & 51.80 & 52.40 & 74.23 & 59.56 & 79.9\% \\
SparseVLM (ICML'25) & 60.39 & 76.25 & 2145 & 77.23 & 51.30 & 61.00 & 53.00 & 76.30 & 49.61 & 81.4\% \\
DivPrune (CVPR'25) & 57.25 & 81.73 & 2158 & 76.12 & 50.22 & 39.00 & 41.70 & 59.32 & 34.58 & 72.5\% \\
DART (EMNLP'25) & 58.82 & 78.60 & 2074 & 78.26 & 48.89 & 44.88 & 42.70 & 57.12 & 33.65 & 72.6\% \\
VisionZip (CVPR'25) & 65.23 & 83.55 & 2147 & 79.04 & 54.90 & 53.08 & 49.40 & 67.16 & 49.55 & 81.1\% \\
MMTok (ICLR'26) & 58.82 & 82.44 & 2218 & 79.47 & 54.04 & 51.04 & 51.30 & 67.33 & 51.37 & 80.4\% \\
PACE (w/o APC) & 65.36 & 82.27 & 2228 & 80.07 & 55.25 & 65.76 & 58.00 & 76.18 & 65.54 & 87.7\% \\
\rowcolor[HTML]{E8F5E9} \textbf{PACE (Ours)} & \textbf{68.37} & \textbf{84.99} & \textbf{2314} & \textbf{83.33} & \textbf{59.08} & \textbf{73.52} & \textbf{70.90} & \textbf{78.42} & \textbf{69.55} & \textbf{93.8\%} \\
\midrule
\rowcolor[HTML]{FAFAFA} \multicolumn{11}{c}{\textit{Retain 5\% $\bar{T}$}} \\
FastV (ECCV'24) & 56.08 & 62.45 & 1995 & 73.54 & 45.88 & 36.52 & 41.40 & 66.21 & 44.17 & 69.6\% \\
SparseVLM (ICML'25) & 56.60 & 65.25 & 2011 & 72.08 & 45.01 & 46.24 & 43.10 & 69.53 & 30.99 & 70.3\% \\
DivPrune (CVPR'25) & 53.46 & 77.66 & 1926 & 72.77 & 45.20 & 28.96 & 29.80 & 41.28 & 22.64 & 62.0\% \\
DART (EMNLP'25) & 52.68 & 71.33 & 1946 & 73.97 & 42.80 & 30.76 & 32.90 & 45.01 & 23.37 & 62.2\% \\
VisionZip (CVPR'25) & 61.57 & 78.65 & 2029 & 75.17 & 48.63 & 40.96 & 38.30 & 55.47 & 29.40 & 70.5\% \\
MMTok (ICLR'26) & 54.12 & 78.70 & 2133 & 75.17 & 47.00 & 32.76 & 36.00 & 50.95 & 28.85 & 67.3\% \\
PACE (w/o APC) & 60.39 & 76.59 & 2117 & 75.95 & 49.26 & 53.00 & 46.10 & 67.32 & 45.38 & 76.9\% \\
\rowcolor[HTML]{E8F5E9} \textbf{PACE (Ours)} & \textbf{63.79} & \textbf{80.70} & \textbf{2245} & \textbf{80.07} & \textbf{56.02} & \textbf{61.84} & \textbf{57.80} & \textbf{71.56} & \textbf{48.94} & \textbf{84.3\%} \\
\midrule
\rowcolor[HTML]{E8EAF6} \multicolumn{11}{c}{\textbf{Dynamic-resolution setting (MinPix = $256 \times 28 \times 28$, MaxPix = $2048 \times 28 \times 28$)}} \\
\midrule
\rowcolor[HTML]{EFEFEF} \textit{Vanilla (100\% Tokens)} & 69.54 & 86.46 & 2308 & 84.28 & 62.75 & 83.92 & 84.30 & 82.94 & 94.73 & 100.0\% \\
\midrule
\rowcolor[HTML]{FAFAFA} \multicolumn{11}{c}{\textit{Retain 20\% $\bar{T}$}} \\
FastV (ECCV'24) & 64.44 & 74.76 & 2157 & 78.18 & 49.69 & 62.60 & 59.90 & 77.49 & 75.93 & 84.9\% \\
SparseVLM (ICML'25) & 64.44 & 79.18 & 2195 & 77.06 & 50.06 & 66.00 & 56.10 & \textbf{80.14} & 72.92 & 85.5\% \\
DivPrune (CVPR'25) & 63.14 & 81.21 & 2076 & 76.37 & 49.39 & 48.16 & 49.20 & 63.93 & 48.56 & 76.5\% \\
DART (EMNLP'25) & 63.01 & 80.72 & 2200 & 77.75 & 48.43 & 44.04 & 52.70 & 65.64 & 50.07 & 77.3\% \\
VisionZip (CVPR'25) & 65.49 & 83.11 & 2190 & 80.58 & 55.61 & 51.00 & 58.30 & 69.18 & 74.76 & 84.6\% \\
MMTok (ICLR'26) & 63.01 & 83.05 & 2209 & 79.12 & 52.36 & 59.60 & 60.20 & 72.00 & 73.81 & 85.2\% \\
PACE (w/o APC) & 65.49 & 82.08 & 2271 & 79.21 & 54.11 & \textbf{70.48} & 64.20 & 78.61 & 82.37 & 90.0\% \\
\rowcolor[HTML]{E8F5E9} \textbf{PACE (Ours)} & \textbf{68.10} & \textbf{83.52} & \textbf{2276} & \textbf{80.93} & \textbf{56.24} & 68.84 & \textbf{72.20} & 78.11 & \textbf{86.35} & \textbf{92.4\%} \\
\midrule
\rowcolor[HTML]{FAFAFA} \multicolumn{11}{c}{\textit{Retain 10\% $\bar{T}$}} \\
FastV (ECCV'24) & 58.56 & 65.45 & 1946 & 73.28 & 44.99 & 45.52 & 45.00 & 70.69 & 58.66 & 73.1\% \\
SparseVLM (ICML'25) & 59.22 & 69.03 & 1930 & 63.65 & 41.45 & 50.00 & 36.20 & \textbf{75.19} & 48.40 & 70.5\% \\
DivPrune (CVPR'25) & 58.69 & 76.61 & 1916 & 72.25 & 43.57 & 36.48 & 37.50 & 50.77 & 33.91 & 66.2\% \\
DART (EMNLP'25) & 56.99 & 73.49 & 2017 & 73.11 & 44.25 & 31.20 & 42.00 & 51.76 & 33.23 & 66.2\% \\
VisionZip (CVPR'25) & 63.14 & \textbf{78.97} & 1959 & 75.00 & 49.03 & 39.56 & 39.20 & 57.67 & 48.55 & 72.1\% \\
MMTok (ICLR'26) & 58.69 & 77.63 & 2020 & 75.08 & 47.31 & 38.96 & 44.80 & 59.07 & 50.55 & 72.3\% \\
PACE (w/o APC) & 63.79 & 74.89 & 2056 & 74.66 & 48.76 & 51.60 & 46.30 & 70.81 & 64.59 & 78.2\% \\
\rowcolor[HTML]{E8F5E9} \textbf{PACE (Ours)} & \textbf{67.32} & 76.86 & \textbf{2112} & \textbf{77.32} & \textbf{51.12} & \textbf{56.40} & \textbf{53.00} & 72.81 & \textbf{68.49} & \textbf{82.3\%} \\
\midrule
\rowcolor[HTML]{FAFAFA} \multicolumn{11}{c}{\textit{Retain 5\% $\bar{T}$}} \\
FastV (ECCV'24) & 51.63 & 47.57 & 1646 & 65.21 & 39.26 & 31.24 & 30.50 & 58.27 & 43.15 & 58.9\% \\
SparseVLM (ICML'25) & 54.90 & 38.89 & 1542 & 47.25 & 34.97 & 29.00 & 19.70 & 63.04 & 29.68 & 52.0\% \\
DivPrune (CVPR'25) & 54.51 & 71.14 & 1767 & 66.58 & 40.58 & 25.32 & 24.50 & 36.85 & 22.23 & 56.5\% \\
DART (EMNLP'25) & 52.94 & 61.60 & \textbf{1846} & 67.44 & 38.48 & 22.24 & 32.70 & 39.55 & 23.03 & 56.2\% \\
VisionZip (CVPR'25) & 61.44 & \textbf{71.22} & 1731 & 69.59 & 41.36 & 26.44 & 23.60 & 43.96 & 28.87 & 59.7\% \\
MMTok (ICLR'26) & 51.37 & 68.85 & 1808 & 66.84 & 40.26 & 23.76 & 31.40 & 43.67 & 28.22 & 58.1\% \\
PACE (w/o APC) & 58.69 & 61.85 & 1736 & 69.07 & 43.55 & 35.84 & 28.50 & 57.51 & 44.28 & 63.9\% \\
\rowcolor[HTML]{E8F5E9} \textbf{PACE (Ours)} & \textbf{63.14} & 64.75 & 1783 & \textbf{72.51} & \textbf{45.36} & \textbf{38.52} & \textbf{32.80} & \textbf{63.36} & \textbf{47.64} & \textbf{68.1\%} \\
\bottomrule
\end{tabular}%
}
\vspace{2mm}
\caption{\textbf{Comprehensive performance on Qwen2.5-VL-7B.}
Results cover the fixed- and dynamic-resolution settings at 20\%, 10\%, and 5\% token-retention budgets.
}
\label{tab:appendix_7b}
\end{table*}

\begin{table*}[htbp]
\centering
\small
\renewcommand{\arraystretch}{1.1}
\resizebox{\textwidth}{!}{%
\begin{tabular}{l|ccccccccc|c}
\toprule
\textbf{Method} & \textbf{RealWorldQA} & \textbf{POPE} & \textbf{MME} & \textbf{MMBench} & \textbf{MMStar} & \textbf{ChartQA}& \textbf{OCRBench}& \textbf{TextVQA} & \textbf{DocVQA} & \textbf{Avg.} \\
& Acc. $\uparrow$ & F1 $\uparrow$ & P+C $\uparrow$ & Acc. $\uparrow$ & Acc. $\uparrow$ & Acc. $\uparrow$ & Acc. $\uparrow$ & Acc. $\uparrow$ & ANLS $\uparrow$ & $\uparrow$ \\
\midrule
\rowcolor[HTML]{E8EAF6} \multicolumn{11}{c}{\textbf{Fixed-resolution setting (MinPix = $2048 \times 28 \times 28$, MaxPix = $2048 \times 28 \times 28$)}} \\
\midrule
\rowcolor[HTML]{EFEFEF} \textit{Vanilla (100\% Tokens)} & 67.52 & 87.40 & 2088 & 77.58 & 55.93 & 84.04 & 72.30 & 78.69 & 93.02 & 100.0\% \\
\midrule
\rowcolor[HTML]{FAFAFA} \multicolumn{11}{c}{\textit{Retain 20\% $\bar{T}$}} \\
FastV (ECCV'24) & 52.03 & 84.44 & 1998 & 73.97 & 50.93 & 74.32 & 60.10 & 74.51 & 74.25 & 89.1\% \\
SparseVLM (ICML'25) & 51.50 & 86.06 & 2044 & 73.88 & 51.31 & 68.24 & 54.90 & 74.47 & 62.74 & 86.5\% \\
DivPrune (CVPR'25) & 48.89 & 84.73 & 1920 & 72.42 & 50.03 & 55.76 & 41.80 & 59.69 & 44.31 & 76.9\% \\
DART (EMNLP'25) & 53.20 & 84.54 & 2043 & 73.45 & 52.12 & 68.44 & 47.30 & 64.01 & 52.16 & 82.8\% \\
VisionZip (CVPR'25) & 53.99 & 86.72 & 2074 & 75.08 & 53.09 & 77.36 & 57.70 & 69.17 & 70.93 & 89.6\% \\
MMTok (ICLR'26) & 53.73 & 86.62 & 2017 & 75.25 & 54.80 & 74.16 & 60.20 & 72.47 & 69.51 & 89.8\% \\
PACE (w/o APC) & 53.86 & 87.08 & 2102 & 75.52 & 53.37 & 79.04 & 60.20 & 72.63 & 75.14 & 91.5\% \\
\rowcolor[HTML]{E8F5E9} \textbf{PACE (Ours)} & \textbf{55.56} & \textbf{87.61} & \textbf{2138} & \textbf{76.98} & \textbf{55.61} & \textbf{82.60} & \textbf{71.40} & \textbf{76.59} & \textbf{82.19} & \textbf{96.3\%} \\
\midrule
\rowcolor[HTML]{FAFAFA} \multicolumn{11}{c}{\textit{Retain 10\% $\bar{T}$}} \\
FastV (ECCV'24) & 48.10 & 78.11 & 1900 & 70.87 & 47.82 & 60.56 & 47.40 & 69.65 & 54.84 & 79.3\% \\
SparseVLM (ICML'25) & 47.58 & 80.43 & 1922 & 69.58 & 47.21 & 43.92 & 40.70 & 68.31 & 40.02 & 74.1\% \\
DivPrune (CVPR'25) & 49.02 & 83.96 & 1893 & 70.10 & 49.01 & 45.68 & 32.70 & 50.44 & 30.87 & 70.5\% \\
DART (EMNLP'25) & 51.50 & 80.43 & 1863 & 69.84 & 46.59 & 51.48 & 34.80 & 50.61 & 32.97 & 71.1\% \\
VisionZip (CVPR'25) & 52.81 & 84.50 & 1932 & 71.64 & 50.59 & 60.84 & 42.50 & 57.23 & 43.32 & 77.9\% \\
MMTok (ICLR'26) & 53.07 & 85.22 & 1916 & 72.68 & 50.82 & 61.00 & 46.50 & 63.48 & 50.62 & 80.5\% \\
PACE (w/o APC) & 50.59 & 84.95 & 1950 & 73.11 & 50.82 & 67.80 & 47.70 & 65.06 & 53.95 & 82.0\% \\
\rowcolor[HTML]{E8F5E9} \textbf{PACE (Ours)} & \textbf{53.73} & \textbf{86.77} & \textbf{2027} & \textbf{74.31} & \textbf{53.95} & \textbf{76.56} & \textbf{60.50} & \textbf{70.33} & \textbf{61.95} & \textbf{88.8\%} \\
\midrule
\rowcolor[HTML]{FAFAFA} \multicolumn{11}{c}{\textit{Retain 5\% $\bar{T}$}} \\
FastV (ECCV'24) & 42.48 & 65.23 & 1742 & 66.67 & 43.01 & 42.56 & 38.60 & 62.16 & 38.56 & 67.6\% \\
SparseVLM (ICML'25) & 40.78 & 64.91 & 1696 & 60.14 & 40.79 & 24.76 & 31.10 & 57.52 & 24.43 & 59.8\% \\
DivPrune (CVPR'25) & 49.02 & 83.96 & 1893 & 66.32 & 46.00 & 45.68 & 32.70 & 50.44 & 22.65 & 68.4\% \\
DART (EMNLP'25) & 48.76 & 73.92 & 1744 & 65.38 & 41.98 & 33.84 & 22.50 & 36.86 & 21.75 & 60.1\% \\
VisionZip (CVPR'25) & 52.81 & 84.50 & \textbf{1932} & 68.21 & 47.04 & 60.84 & 42.50 & 57.23 & 26.45 & 74.6\% \\
MMTok (ICLR'26) & \textbf{53.07} & \textbf{85.22} & 1919 & 67.87 & 47.43 & 61.00 & \textbf{46.50} & \textbf{63.48} & 31.29 & 76.8\% \\
PACE (w/o APC) & 47.06 & 79.96 & 1804 & 70.36 & 47.65 & 52.84 & 37.10 & 53.94 & 34.06 & 71.4\% \\
\rowcolor[HTML]{E8F5E9} \textbf{PACE (Ours)} & 50.72 & 83.66 & 1896 & \textbf{73.54} & \textbf{51.33} & \textbf{63.04} & 46.10 & 61.33 & \textbf{40.71} & \textbf{78.7\%} \\
\midrule
\rowcolor[HTML]{E8EAF6} \multicolumn{11}{c}{\textbf{Dynamic-resolution setting (MinPix = $256 \times 28 \times 28$, MaxPix = $2048 \times 28 \times 28$)}} \\
\midrule
\rowcolor[HTML]{EFEFEF} \textit{Vanilla (100\% Tokens)} & 59.74 & 86.47 & 2154 & 78.01 & 55.67 & 83.36 & 77.90 & 78.75 & 92.89 & 100.0\% \\
\midrule
\rowcolor[HTML]{FAFAFA} \multicolumn{11}{c}{\textit{Retain 20\% $\bar{T}$}} \\
FastV (ECCV'24) & 54.12 & 78.73 & 2002 & 71.56 & 46.91 & 64.76 & 54.50 & 73.29 & 72.84 & 85.5\% \\
SparseVLM (ICML'25) & 54.51 & 81.63 & 2005 & 70.18 & 46.07 & 56.08 & 46.20 & \textbf{74.41} & 61.74 & 82.1\% \\
DivPrune (CVPR'25) & 53.20 & 79.90 & 1877 & 69.15 & 45.42 & 51.40 & 43.40 & 56.52 & 43.77 & 75.0\% \\
DART (EMNLP'25) & 55.29 & 80.52 & 1926 & 70.10 & 46.39 & 53.00 & 45.30 & 55.77 & 51.40 & 77.3\% \\
VisionZip (CVPR'25) & \textbf{57.25} & 82.80 & 1905 & 71.90 & 49.21 & 65.48 & 51.30 & 64.22 & 70.17 & 84.7\% \\
MMTok (ICLR'26) & 55.42 & 82.23 & 2053 & 72.59 & 48.22 & 66.16 & 56.30 & 67.56 & 69.23 & 86.1\% \\
PACE (w/o APC) & 55.95 & 83.43 & 1948 & 72.51 & 49.97 & 70.48 & 56.90 & 71.49 & 74.49 & 88.0\% \\
\rowcolor[HTML]{E8F5E9} \textbf{PACE (Ours)} & 56.86 & \textbf{84.29} & \textbf{2065} & \textbf{75.69} & \textbf{52.83} & \textbf{71.20} & \textbf{64.80} & 71.77 & \textbf{81.74} & \textbf{92.0\%} \\
\midrule
\rowcolor[HTML]{FAFAFA} \multicolumn{11}{c}{\textit{Retain 10\% $\bar{T}$}} \\
FastV (ECCV'24) & 49.93 & 68.80 & 1834 & 65.46 & 41.38 & 48.32 & 41.70 & 67.52 & \textbf{54.09} & 73.5\% \\
SparseVLM (ICML'25) & 47.97 & 72.65 & 1801 & 55.84 & 39.68 & 28.48 & 27.50 & \textbf{69.37} & 39.16 & 65.6\% \\
DivPrune (CVPR'25) & 51.11 & 75.54 & 1771 & 64.86 & 41.52 & 42.20 & 32.20 & 46.53 & 30.19 & 66.3\% \\
DART (EMNLP'25) & 53.99 & 72.37 & 1763 & 64.08 & 41.64 & 37.48 & 30.50 & 41.88 & 32.39 & 65.0\% \\
VisionZip (CVPR'25) & \textbf{55.82} & 77.53 & 1722 & 67.09 & 44.50 & 47.68 & 32.80 & 48.28 & 42.41 & 70.6\% \\
MMTok (ICLR'26) & 53.73 & \textbf{79.05} & \textbf{1921} & 68.47 & 43.58 & 47.60 & 40.90 & 56.51 & 49.81 & 74.6\% \\
PACE (w/o APC) & 52.81 & 77.79 & 1795 & 68.21 & 46.30 & 53.92 & 38.80 & 62.02 & 52.97 & 75.8\% \\
\rowcolor[HTML]{E8F5E9} \textbf{PACE (Ours)} & 52.42 & 78.00 & 1855 & \textbf{71.05} & \textbf{48.46} & \textbf{56.44} & \textbf{43.00} & 63.52 & 52.42 & \textbf{78.0\%} \\
\midrule
\rowcolor[HTML]{FAFAFA} \multicolumn{11}{c}{\textit{Retain 5\% $\bar{T}$}} \\
FastV (ECCV'24) & 45.23 & 54.81 & 1630 & 54.04 & 37.36 & 31.08 & 26.80 & \textbf{58.50} & 37.71 & 59.8\% \\
SparseVLM (ICML'25) & 43.79 & 51.84 & 1533 & 38.66 & 35.52 & 16.16 & 13.00 & 57.54 & 23.82 & 50.3\% \\
DivPrune (CVPR'25) & 48.63 & 71.11 & 1577 & 57.82 & 36.92 & 33.52 & 21.30 & 36.47 & 21.94 & 57.2\% \\
DART (EMNLP'25) & 50.33 & 62.27 & 1548 & 55.93 & 37.34 & 25.76 & 22.10 & 30.35 & 21.09 & 54.2\% \\
VisionZip (CVPR'25) & \textbf{51.90} & 69.90 & 1557 & 59.62 & 39.89 & 32.96 & 18.90 & 35.46 & 25.93 & 58.3\% \\
MMTok (ICLR'26) & 48.89 & \textbf{73.17} & \textbf{1709} & 59.36 & 39.21 & 23.76 & \textbf{28.50} & 42.57 & 30.47 & 60.5\% \\
PACE (w/o APC) & 50.98 & 64.11 & 1541 & 60.48 & \textbf{41.26} & \textbf{38.80} & 22.20 & 48.90 & 33.49 & 61.8\% \\
\rowcolor[HTML]{E8F5E9} \textbf{PACE (Ours)} & 50.72 & 65.31 & 1652 & \textbf{62.03} & 41.10 & 37.52 & 26.70 & 52.05 & \textbf{39.57} & \textbf{64.3\%} \\
\bottomrule
\end{tabular}%
}
\vspace{2mm}
\caption{\textbf{Comprehensive performance on Qwen2.5-VL-3B.}
Results cover the fixed- and dynamic-resolution settings at 20\%, 10\%, and 5\% token-retention budgets.
}
\label{tab:appendix_3b}
\end{table*}

\subsection{TTFT Profiling Across Retention Budgets}
\label{sec:ttft_budget_sweep}

Table~\ref{tab:ttft_budget_sweep} extends the main 10\% measurement to milder retention budgets. TTFT is the complete pre-generation latency: Vanilla includes vision encoding and LLM prefill, while PACE additionally includes the Shallow Feature Preview and adaptive resizing. The isolated encoder and prefill columns exclude these APC overheads. Autoregressive decoding is not included because it is not accelerated by PACE.

\begin{table*}[htbp]
\centering
\small
\setlength{\tabcolsep}{4pt}
\renewcommand{\arraystretch}{1.12}
\resizebox{\textwidth}{!}{%
\begin{tabular}{cc|ccc}
\toprule
\textbf{Retention} & \textbf{Dataset} & \textbf{Encoder: Vanilla / PACE (Spd.)} & \textbf{Prefill: Vanilla / PACE (Spd.)} & \textbf{TTFT incl. APC: Vanilla / PACE (Spd.)} \\
\midrule
\multirow{2}{*}{80\%} & DocVQA & 144.77 / 109.07 ($1.33\times$) & 209.35 / 168.63 ($1.24\times$) & 354.12 / 321.57 ($1.10\times$) \\
& TextVQA & 156.27 / 117.33 ($1.33\times$) & 220.02 / 174.56 ($1.26\times$) & 376.29 / 339.98 ($1.11\times$) \\
\midrule
\multirow{2}{*}{50\%} & DocVQA & 144.69 / 66.28 ($2.18\times$) & 207.84 / 105.81 ($1.96\times$) & 352.54 / 210.30 ($1.68\times$) \\
& TextVQA & 155.24 / 70.17 ($2.21\times$) & 220.40 / 114.70 ($1.92\times$) & 375.64 / 224.88 ($1.67\times$) \\
\midrule
\multirow{2}{*}{20\%} & DocVQA & 144.73 / 51.97 ($2.78\times$) & 207.72 / 50.92 ($4.08\times$) & 352.45 / 138.53 ($2.54\times$) \\
& TextVQA & 155.20 / 49.46 ($3.14\times$) & 220.26 / 51.32 ($4.29\times$) & 375.46 / 137.34 ($2.73\times$) \\
\midrule
\rowcolor[HTML]{E8F5E9}
\multirow{2}{*}{\textbf{10\%}} & \textbf{DocVQA} & \textbf{143.41 / 50.74 ($2.83\times$)} & \textbf{210.41 / 32.70 ($6.43\times$)} & \textbf{353.82 / 117.54 ($3.01\times$)} \\
\rowcolor[HTML]{E8F5E9}
& \textbf{TextVQA} & \textbf{154.28 / 48.20 ($3.20\times$)} & \textbf{223.68 / 32.69 ($6.84\times$)} & \textbf{377.96 / 116.03 ($3.26\times$)} \\
\bottomrule
\end{tabular}%
}
\caption{\textbf{Latency across visual-token retention budgets on Qwen2.5-VL-7B.}
 Average per-sample milliseconds are measured under the fixed-resolution setting on one RTX 4090. At 80\% retention (only 20\% token reduction), PACE already improves TTFT by about $1.10\times$; its benefit grows as more redundant computation is removed.}
\label{tab:ttft_budget_sweep}
\end{table*}

The average TTFT speedup across DocVQA and TextVQA is $1.10\times$, $1.67\times$, $2.64\times$, and $3.13\times$ at 80\%, 50\%, 20\%, and 10\% retention, respectively. The modest 80\% gain reflects that preview overhead is nearly fixed while relatively little encoder and prefill computation is removed; the overhead is progressively amortized at stricter budgets.

\paragraph{Dataset-wise 10\% profiling.}
\label{sec:appendix_dataset_efficiency}
We further report the number of evaluated samples and average per-sample latency over all nine benchmarks. As shown in Table~\ref{tab:appendix_dataset_efficiency}, PACE consistently reduces both encoder-side and LLM-side latency. On average, it reduces encoder latency from 152.41 ms to 46.10 ms and prefill latency from 222.16 ms to 31.63 ms, corresponding to $3.32\times$ and $7.02\times$ speedups. Including APC and resizing, average TTFT decreases from 374.58 ms to 112.08 ms ($3.34\times$).

\begin{table*}[t]
\centering
\small
\setlength{\tabcolsep}{3.2pt}
\renewcommand{\arraystretch}{1.15}
\resizebox{\textwidth}{!}{%
\begin{tabular}{l r | cc c | cc c | cc c}
\toprule
\multirow{2}{*}{\textbf{Benchmark}} 
& \multirow{2}{*}{\textbf{\# Samples}} 
& \multicolumn{3}{c|}{\textbf{Encoder Time (ms)}} 
& \multicolumn{3}{c|}{\textbf{Prefill Time (ms)}} 
& \multicolumn{3}{c}{\textbf{TTFT incl. APC (ms)}} \\
\cmidrule(lr){3-5}
\cmidrule(lr){6-8}
\cmidrule(lr){9-11}
& 
& \textbf{Van.} & \textbf{PACE} & \textbf{Spd.}
& \textbf{Van.} & \textbf{PACE} & \textbf{Spd.}
& \textbf{Van.} & \textbf{PACE} & \textbf{Spd.} \\
\midrule
RealWorldQA 
& 765 
& 152.33 & 44.83 & $3.40\times$ 
& 222.86 & 31.59 & $7.05\times$ 
& 375.19 & 110.50 & $3.40\times$ \\

POPE 
& 9,000 
& 154.90 & 47.71 & $3.25\times$ 
& 224.60 & 31.60 & $7.11\times$ 
& 379.50 & 114.37 & $3.32\times$ \\

MME 
& 2,374 
& 152.28 & 46.85 & $3.25\times$ 
& 221.92 & 31.58 & $7.03\times$ 
& 374.20 & 112.78 & $3.32\times$ \\

MMBench 
& 4,329 
& 154.02 & 43.23 & $3.56\times$ 
& 224.17 & 31.63 & $7.09\times$ 
& 378.19 & 108.91 & $3.47\times$ \\

MMStar 
& 1,500 
& 153.76 & 44.04 & $3.49\times$ 
& 223.87 & 31.66 & $7.07\times$ 
& 377.63 & 109.87 & $3.44\times$ \\

ChartQA 
& 2,500 
& 152.80 & 45.03 & $3.39\times$ 
& 223.33 & 31.57 & $7.07\times$ 
& 376.13 & 110.91 & $3.39\times$ \\

OCRBench 
& 1,000 
& 153.13 & 44.50 & $3.44\times$ 
& 223.06 & 31.76 & $7.02\times$ 
& 376.20 & 110.38 & $3.41\times$ \\

TextVQA 
& 5,000 
& 154.70 & 48.10 & $3.22\times$ 
& 224.53 & 31.68 & $7.09\times$ 
& 379.23 & 114.91 & $3.30\times$ \\

DocVQA 
& 5,349 
& 143.79 & 50.58 & $2.84\times$ 
& 211.16 & 31.58 & $6.69\times$ 
& 354.94 & 116.09 & $3.06\times$ \\
\midrule
\rowcolor[HTML]{E8F5E9}
\textbf{Macro Avg.} 
& -- 
& \textbf{152.41} & \textbf{46.10} & $\mathbf{3.32\times}$ 
& \textbf{222.16} & \textbf{31.63} & $\mathbf{7.02\times}$ 
& \textbf{374.58} & \textbf{112.08} & $\mathbf{3.34\times}$ \\
\bottomrule
\end{tabular}%
}
\vspace{1mm}
\caption{
\textbf{Dataset-wise latency profiling on Qwen2.5-VL-7B at 10\% retention.}
 Average per-sample latency is reported in milliseconds under the fixed-resolution setting. Isolated stage times exclude APC, whereas PACE TTFT includes its preview and resizing overhead.
}
\label{tab:appendix_dataset_efficiency}
\end{table*}

\subsection{Full DDAE Extraction-Depth Latency}
\label{sec:depth_latency_full}

Table~\ref{tab:depth_latency_full} reports the complete latency counterpart to the quality ablation in Table~\ref{tab:ddae_depth_ablation}. Measurements use the same 400 samples under two fixed-resolution settings. Delaying DDAE monotonically erodes prefill acceleration: moving from layer 2 to layer 24 reduces speedup from $2.11\times$ to $1.09\times$ at 2048 patch units and from $2.32\times$ to $1.10\times$ at 4096. Decode and time per output token remain approximately unchanged ($\approx1.00\times$), confirming that PACE's measured acceleration should be attributed to TTFT rather than the full generation process.

\begin{table}[ht]
\centering
\small
\setlength{\tabcolsep}{3.5pt}
\renewcommand{\arraystretch}{1.1}
\resizebox{\columnwidth}{!}{%
\begin{tabular}{c c|ccc}
\toprule
\textbf{Fixed-resolution setting} & \textbf{Depth} & \textbf{w/o DDAE (ms)} & \textbf{DDAE (ms)} & \textbf{Spd.} \\
\midrule
\multirow{4}{*}{$2048\times28\times28$}
& 2 & 74.52 & 35.26 & $2.11\times$ \\
& 8 & 74.52 & 42.51 & $1.75\times$ \\
& 16 & 74.52 & 55.32 & $1.35\times$ \\
& 24 & 74.52 & 68.12 & $1.09\times$ \\
\midrule
\multirow{4}{*}{$4096\times28\times28$}
& 2 & 134.65 & 58.01 & $2.32\times$ \\
& 8 & 134.65 & 75.57 & $1.78\times$ \\
& 16 & 134.65 & 99.11 & $1.36\times$ \\
& 24 & 134.65 & 122.77 & $1.10\times$ \\
\bottomrule
\end{tabular}%
}
\caption{\textbf{LLM prefill latency across DDAE extraction depths.}
 ``w/o DDAE'' processes the full visual sequence throughout prefill. Earlier extraction leaves fewer full-sequence layers and therefore yields greater acceleration.}
\label{tab:depth_latency_full}
\end{table}

\section{Detailed Computational Complexity Analysis}
\label{sec:detailed_complexity}

We analyze PACE using \textbf{Qwen2.5-VL} \citep{bai2025qwen25vltechnicalreport}, distinguishing encoder patches from the post-merger visual tokens passed to the LLM.

\subsection{Architecture and Stage-wise Bottlenecks of Qwen2.5-VL}
An $H\times W$ image produces $N=HW/P^2$ encoder patches of dimension $D_v$. Each of the $L_v$ ViT blocks costs $\mathcal{O}(ND_v^2+N^2D_v)$. A spatial merger then produces $N_{vis}$ visual tokens in the LLM dimension $D_l$. With $T$ prompt tokens, Vanilla has prefill length $S_0=T+N_{vis}$ and per-layer cost $\mathcal{O}(S_0D_l^2+S_0^2D_l)$. Autoregressive decoding determines TPOT and is unchanged by PACE.

\subsection{Theoretical Complexity Reduction}
\label{sec:theoretical_complexity_reduction}

Let $p_1\in(0,1]$ be APC's retained image-area ratio. The full encoder processes $N_{enc}\approx p_1N$ patches, and the merger outputs $N_{vis}^{c}\approx p_1N_{vis}$ tokens. Because the preview uses $K=1$ block at the original resolution, the vision-side cost is
\begin{equation}
\begin{aligned}
 \mathcal{F}_{\mathrm{ViT}}^{\mathrm{PACE}}
 &= \mathcal{C}_{\mathrm{prev}}+\mathcal{C}_{\mathrm{enc}},\\
 \mathcal{C}_{\mathrm{prev}}
 &= \mathcal{O}(KND_v^2+KN^2D_v),\quad K=1,\\
 \mathcal{C}_{\mathrm{enc}}
 &= \mathcal{O}(L_vp_1ND_v^2+L_vp_1^2N^2D_v).
\end{aligned}
\end{equation}
Let $p_2\in(0,1]$ be DDAE's retention after condensation and $\mathcal{B}=p_1p_2$ the final ratio relative to Vanilla. Then
\begin{equation}
 N_{keep}=p_2N_{vis}^{c}\approx \mathcal{B}N_{vis}.
\end{equation}
Define $S_{pre}=T+N_{vis}^{c}$ and $S_{post}=T+N_{keep}$. Extracting after layer $L_{ext}$ in an $L_l$-layer LLM gives
\begin{equation}
\begin{aligned}
 \mathcal{F}_{\mathrm{pre}}^{\mathrm{PACE}}
 ={}&\mathcal{O}(L_{ext}S_{pre}D_l^2
       +L_{ext}S_{pre}^2D_l)\\
 &+\mathcal{O}((L_l-L_{ext})S_{post}D_l^2\\
 &\hspace{15mm}+(L_l-L_{ext})S_{post}^2D_l).
\end{aligned}
\end{equation}
APC therefore reduces full-encoder computation, whereas earlier DDAE extraction leaves fewer LLM layers operating on $S_{pre}$.

\section{Qualitative Results}
\label{sec:qualitative_results}

\begin{figure*}[t]
    \centering
    \includegraphics[width=0.98\textwidth]{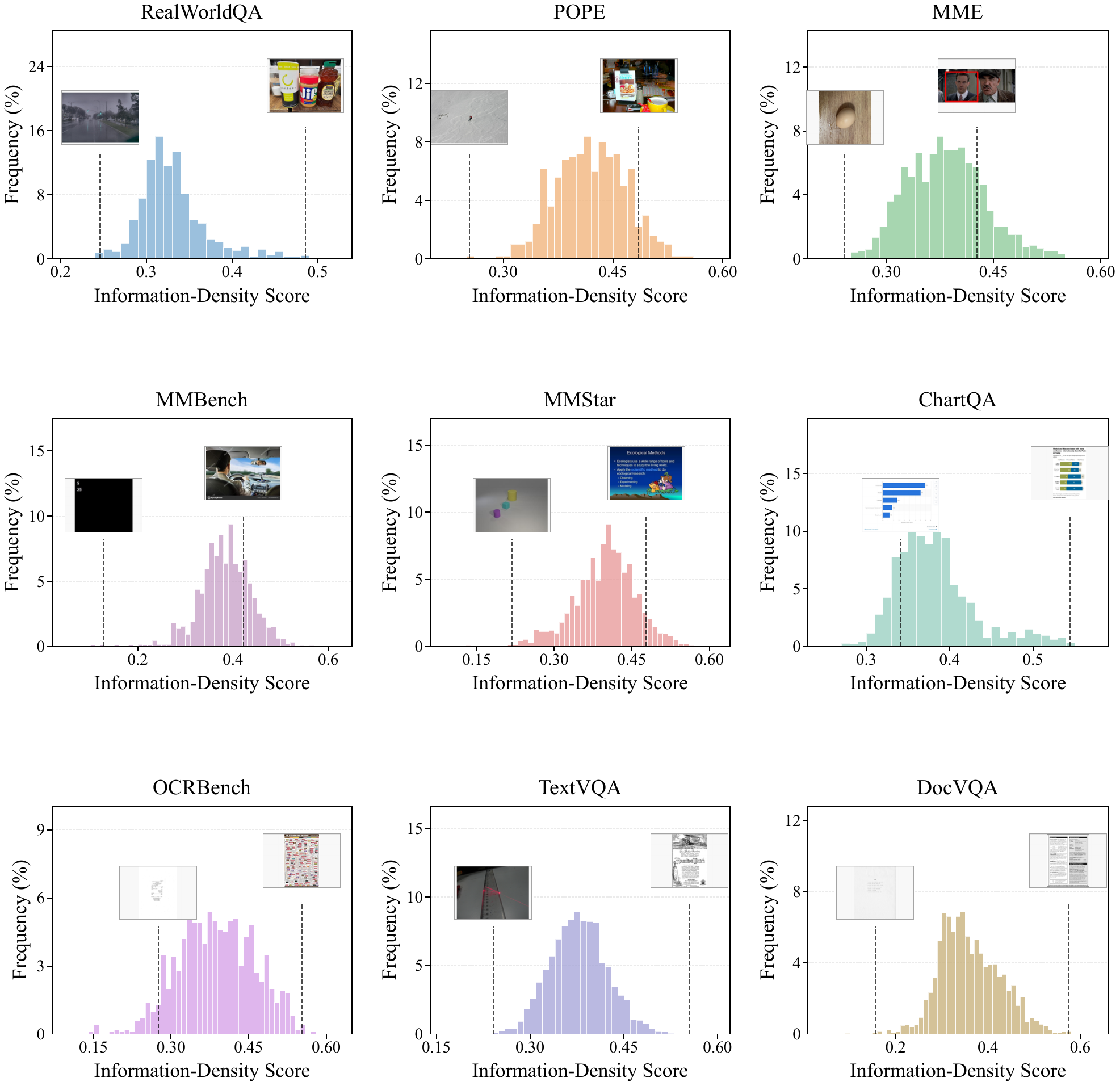}
    \caption{
    \textbf{Information density score distributions across benchmarks.}
    Histograms and examples contrast APC's information density scores across nine benchmarks.
    }
    \label{fig:information_density_distribution}
\end{figure*}

To better understand the behavior of APC across heterogeneous visual domains, we visualize the distribution of the information density score across all evaluation benchmarks in Figure~\ref{fig:information_density_distribution}, alongside representative samples at varying density levels.

As shown in Figure~\ref{fig:information_density_distribution}, the information density score varies noticeably across benchmarks. General visual-understanding datasets, such as RealWorldQA, POPE, MME, and MMBench, exhibit relatively concentrated distributions, indicating that many samples harbor moderate visual redundancy. In contrast, detail-sensitive datasets, including ChartQA, OCRBench, TextVQA, and DocVQA, display broader or heavier high-density regions, reflecting the presence of fine-grained textual, structural, or layout information. These samples are highly vulnerable to uniform downsampling or aggressive post-encoder pruning, as small visual elements frequently contain task-critical evidence.

This qualitative evidence complements the quantitative results. A single fixed-resolution policy cannot consistently accommodate the diverse visual characteristics across benchmarks: low-density samples can be aggressively condensed, whereas high-density samples mandate higher pixel budgets to preserve local details. By estimating information density before full visual encoding, APC dynamically adjusts the input resolution according to the intrinsic complexity of each image, preserving holistic structures while retaining fine-grained evidence under strict token budgets.

\section{Future Work}
\label{sec:future_works}

The empirical success of PACE establishes pre-encoder pixel allocation as a promising frontier for efficient VLM inference. Future research can advance this paradigm across three dimensions. First, an autoregressive recovery mechanism could dynamically fetch high-resolution crops when fine-grained evidence is insufficient, unlocking more aggressive compression. Second, pixel allocation can be advanced to a query-aware regime---conditioned on textual prompts rather than purely intrinsic visual redundancy---using denoised frameworks like VTC-Bench \citep{liao2025we} for precise evaluation. Finally, integrating PACE with localized patching paradigms, such as density-based boundary allocation \citep{choudhury2025accelerating, liuone}, and model-quantization techniques \citep{lin2024awq, wang2026twinquant} could further accelerate end-to-end inference by reducing both visual-token and model-weight computation.

\newpage
\section{LLM Usage Disclosure Statement}
During manuscript preparation, we used LLMs to assist with language editing and debugging experimental code. The authors independently developed the ideas, experimental design, implementation, analysis, and core technical content. We reviewed and verified all LLM-assisted content and take full responsibility for the submission's originality, scientific integrity, and technical accuracy.

\end{document}